\documentclass{article}

\usepackage{microtype}
\usepackage{graphicx}
\usepackage{subfigure}
\usepackage{booktabs} % for professional tables
\usepackage{tabularx}
\usepackage[table,xcdraw]{xcolor}

\usepackage{hyperref}

\definecolor{deemph}{gray}{0.35}
\newcommand{\gc}[1]{\textcolor{deemph}{#1}}

\usepackage[accepted]{icml2024}

\usepackage{amsmath}
\usepackage{amssymb}
\usepackage{mathtools}
\usepackage{amsthm}
\usepackage{bbm}

\newcommand{\W}{\mathbf{W}}
\newcommand{\F}{\mathbf{F}}
\newcommand{\M}{\mathbf{M}}

\newcommand{\X}{\mathbf{X}}
\newcommand{\Y}{\mathbf{Y}}

\newcommand{\R}{\mathbb{R}}

\definecolor{cgray}{HTML}{C0C0C0}

\usepackage[capitalize,noabbrev]{cleveref}

\theoremstyle{plain}

\theoremstyle{definition}

\theoremstyle{remark}

\usepackage{listings}
\usepackage{multirow}
\usepackage{graphicx}
\usepackage{booktabs}
\usepackage{bbding}
\usepackage{makecell}

\usepackage[textsize=tiny]{todonotes}
\newcommand\blfootnote[1]{%
\begingroup
\renewcommand\thefootnote{}\footnote{#1}%
\addtocounter{footnote}{-1}%
\endgroup
}

\icmltitlerunning{SPP: Sparsity-Preserved Parameter-Efficient Fine-Tuning for Large Language Models}

\begin{document}

\twocolumn[
\icmltitle{SPP: Sparsity-Preserved Parameter-Efficient Fine-Tuning\\ for Large Language Models}



\icmlsetsymbol{equal}{*}

\begin{icmlauthorlist}
\icmlauthor{Xudong Lu}{equal,sch1}
\icmlauthor{Aojun Zhou}{equal,sch1}
\icmlauthor{Yuhui Xu}{equal,sch2}
\icmlauthor{Renrui Zhang}{sch1,comp}
\icmlauthor{Peng Gao}{comp}
\icmlauthor{Hongsheng Li$^\dagger$}{sch1,comp2}
\vskip 0.2em
$^1$Multimedia Laboratory (MMLab), The Chinese University of Hong Kong\quad $^2$Salesforce AI Research\\ $^3$Shanghai Artificial Intelligence Laboratory\quad $^4$CPII under InnoHK\\
\vskip 0.2em
\texttt{\{luxudong@link,hsli@ee\}.cuhk.edu.hk}\quad \texttt{\{aojunzhou,xyh6666\}@gmail.com}
\end{icmlauthorlist}

% \icmlaffiliation{comp}{Shanghai Artificial Intelligence Laboratory}
% \icmlaffiliation{comp2}{CPII under InnoHK}
% \icmlaffiliation{sch1}{Multimedia Laboratory (MMLab), The Chinese University of Hong Kong}
% \icmlaffiliation{sch2}{Salesforce AI Research}

% \icmlcorrespondingauthor{Hongsheng Li}{hsli@ee.cuhk.edu.hk}

\icmlkeywords{Machine Learning, ICML}

\vskip 0.3in
]




\begin{abstract}
Large Language Models (LLMs) have become pivotal in advancing the field of artificial intelligence, yet their immense sizes pose significant challenges for both fine-tuning and deployment. Current post-training pruning methods, while reducing the sizes of LLMs, often fail to maintain their original performance. To address these challenges, this paper introduces SPP, a \textbf{S}parsity-\textbf{P}reserved \textbf{P}arameter-efficient fine-tuning method. Different from existing post-training pruning approaches that struggle with performance retention, SPP proposes to employ lightweight learnable column and row matrices to optimize sparse LLM weights, \textit{keeping the structure and sparsity of pruned pre-trained models intact}. By element-wise multiplication and residual addition, SPP ensures the consistency of model sparsity pattern and ratio during both training and weight-merging processes. We demonstrate the effectiveness of SPP by applying it to the LLaMA and LLaMA-2 model families with recent post-training pruning methods. Our results show that SPP significantly enhances the performance of models with different sparsity patterns (i.e. unstructured and N:M sparsity), especially for those with high sparsity ratios (e.g. 75\%), making it a promising solution for the efficient fine-tuning of sparse LLMs. Code will be made available at \url{https://github.com/Lucky-Lance/SPP}.
\end{abstract}

\blfootnote{$^*$Equal contribution\quad $^\dagger$Corresponding author}
\vspace{-3em}
\section{Introduction}
\label{sec: intro}

\begin{figure}[h]
    \centering
    \vspace{0.5em}
    \includegraphics[width=\linewidth]{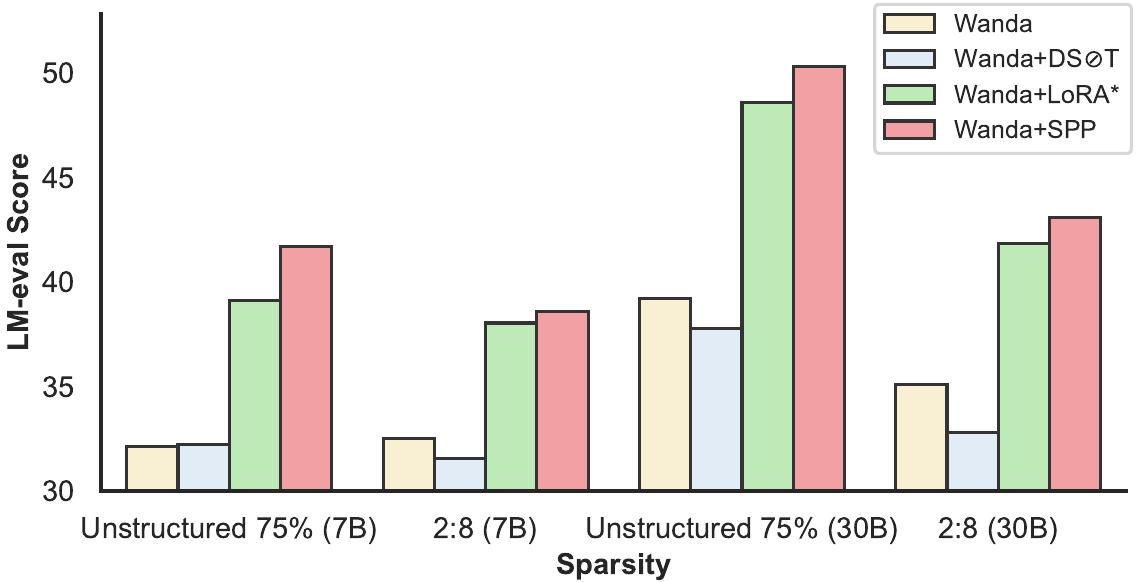}
    \vspace{-0.5em}
    \caption{We use LLaMA 7B/30B models at 75\% sparsity and test zero-shot accuracies on 7 benchmarks of LM-eval~\cite{gao2021framework} to compare different pruning methods. The results of Wanda, Wanda+DS$\oslash$T, Wanda+LoRA*, and Wanda+SPP are visualized. The first two approaches are post-training pruning schemes, LoRA* denotes applying the original Wanda pruning masks to sparse the dense model after LoRA training. Our method achieves overall best results. More experiment details are illustrated in Sec.~\ref{experiments}.}
    \label{fig:bar_plot}
    \vspace{-1.5em}
\end{figure}

Large language models (LLMs)~\cite{brown2020language, OpenAI2023GPT4TR, anil2023palm} have recently shown impressive success in various complex tasks~\cite{wei2022emergent, zhou2024solving}. However, these models are usually characterized by an extensive number of learnable parameters, ranging from several billions to around a hundred billions~\cite{touvron2023llama1,touvron2023llama2}, as exemplified by GPT-4. This enormity makes LLMs cumbersome to be fine-tuned for different scenarios and challenging to deploy on various edge devices.

To reduce the parameters of \textit{pre-trained} large language models without the demanding retraining phase, different post-training pruning~\cite{Frantar2023SparseGPTML} approaches have been presented. SparseGPT, as outlined in~\cite{Frantar2023SparseGPTML}, focuses on minimizing the squared errors between pruned and original dense models in a layer-by-layer manner. Wanda~\cite{sun2023simple} incorporates both weight magnitude and input activation as metrics for identifying unimportant parameters. Despite their success in language model pruning, these methods often fail to maintain the performance of pre-trained models at even moderate sparsity levels (e.g., 50\%)~\cite{jaiswal2023compressing}.

In contrast, pruning methods for smaller-scale deep models -- those with fewer than 200 million parameters, such as ResNet50 ~\cite{he2016deep} and BERT-based models~\cite{devlin2018bert} -- have realized high sparsity ratios (e.g., \textgreater 80\%) only with negligible performance drop~\cite{evci2020rigging,lin2020dynamic,zhou2021learning}. The success of these models is largely attributed to the role of \textbf{retraining} during their pruning process~\cite{liu2018rethinking}. However, compared
with training dense neural networks, it will take much longer to train the sparse models to achieve the same performance. For example, 5$\times$
more training time is needed in RigL~\cite{evci2020rigging}. These methods heavily rely on back-propagation with full parameters, which is prohibitively expensive for LLMs. This observation raises a crucial question: \textit{Can we introduce an efficient retraining stage for the pruned LLMs?}

%Finetuning the full parameters of pruned large language models, similar to the existing static or dynamic sparse training approaches used on models like ResNets and BERTs ~\cite{evci2020rigging, lin2020dynamic}, poses a challenge, particularly in the context of memory consumption. There have been various community developments in parameter-efficient fine-tuning methods, such as LoRA, to enhance the efficiency of fine-tuning large-scale models. However, these techniques, including LoRA, are typically designed for dense models and can not preserve the intrinsic sparse characteristics of pruned models.

In response to these challenges, we propose a novel, plug-and-play method, \textbf{SPP}, a \textbf{S}parsity-\textbf{P}reserved \textbf{P}aramater-efficient fine-tuning method designed to effectively retrain or fine-tune sparse LLMs after post-training pruning (e.g. SparseGPT~\cite{Frantar2023SparseGPTML}, Wanda~\cite{sun2023simple}, etc), thereby enhancing their performances. Inspired by the Low-Rank Adaptation (LoRA) method for dense large language models~\cite{hu2021lora}, \textbf{SPP} consists of two phases, training and weight-merging. More specifically, we introduce two sets of lightweight learnable parameters to the sparse matrices of each linear layer. During training and weight-merging phases, these learnable parameters are multiplied with the original frozen post-training pruned weights, achieving the effect of \textit{exactly maintaining the sparse pattern and ratio} throughout all the processes. %\textbf{SPP} introduces two independent learnable column vectors and row matrices to optimize the sparse weights, which can keep the post-training pruned pre-trained weights frozen and \textit{all weights can exactly maintain the sparse pattern and ratio.} \aj{need to introduce the method details}

SPP is easy to implement %\aj{with only a single easy-to-tune hyperparameter} 
and can be applied to a wide range of post-training pruning methods and LLMs {with various sparsity ratios and patterns (unstructured and N:M sparsity)}. We evaluate SPP on the LLaMA~\cite{touvron2023llama1} and LLaMA-2~\cite{touvron2023llama2} model families with two recent LLM post-training pruning methods, i.e., SparseGPT~\cite{Frantar2023SparseGPTML} and Wanda~\cite{sun2023simple}. To illustrate the effectiveness of SPP, zero-shot evaluation results of LLaMA 7B/30B models with 75\% sparsity ratio are shown in Fig.~\ref{fig:bar_plot}. 

The main contributions of this paper are summarized in three key aspects:

\textbf{(1)} We investigate model pruning methods in the era of LLMs and present a novel parameter-efficient fine-tuning algorithm, SPP, which can maintain model structure and sparsity during both training and weight-merging phases. 

\textbf{(2)} Extensive experiments on different post-training pruned LLMs with various sparsity patterns 
and ratios show the effectiveness of SPP for the efficient training of sparse LLMs. 

\textbf{(3)} To the best of our knowledge, this study is the first to systematically explore integrating efficient retraining with advanced post-training pruning methods for LLMs.

%In contrast, previous pruning methods for smaller-scale deep models (less than 200M parameters, such as ResNet50~\cite{he2016deep} and BERT-base~\cite{devlin2018bert}) have achieved higher sparsity ratios compared to their dense counterparts. The key factor behind their success is the role of \textbf{retraining} in the pruning process. Therefore, This leads to a pivotal question: \textit{How about introducing the training stage for Pruned LLMs ?}

%It is difficult to finetune the full parameters akin to the existing static/dynamic sparse training~\cite{evci2020rigging, lin2020dynamic} on ResNets and Berts. 
%Regardless of the model sparsification, communities develop various parameter-efficient fine-tuning methods (e.g, Lora) to efficiently finetune large-scale models. However, Lora  , such techniques only work for dense models and destroy the inherent sparse characteristics of pruned models. 

%To overcome the difficulty of training for pruned models, we propose a plug-and-play efficient method to finetune the pruned models (e.g., SparseGPT~\cite{Frantar2023SparseGPTML}, Wanda~\cite{sun2023simple}) and improve the performance. Specifically, 

% \newpage
\section{Related Work}
\label{related_work}

\textbf{Traditional Model Pruning:} Pruning of Deep Neural Networks (DNNs) is a promising direction to compress and accelerate deep learning models~\cite{hoefler2021sparsity}. There are mainly two types of techniques to obtain a sparse neural network, iterative pruning-retraining framework, and dynamic sparse training ones. The iterative pruning-retraining framework first finds the unimportant connections (masks), thereby removing the corresponding weights, and then retrains the fixed sparse network to recover its performance. Typical methods include iterative pruning~\cite{Han2015LearningBW}. Later, the lottery ticket hypothesis ~\cite{Frantar2023SparseGPTML} shows that the sparse sub-networks (winning tickets) can be trained from scratch with the same dense initialization while the winning tickets are discovered by dense training. Dynamic sparse training methods~\cite{mocanu2018scalable} start from a randomly sparsified network, then subsequently prune and grow connections during training. These methods can be applied end-to-end within the network training stage and have achieved promising results. Recently proposed state-of-the-art method STR~\citep{kusupati2020soft} introduces learnable pruning thresholds to obtain a non-uniform sparse network. RigL~\citep{evci2020rigging} uses the magnitude-based method to prune and the periodic dense gradients to regrow connections. {However, it is imperative to note that these methods heavily rely on back-propagation with full parameters, which is prohibitively expensive for LLMs.}

%previous pruning methods~\cite{evci2020rigging, lin2020dynamic} required much more training overhead than the dense models training, which are not affordable for large language models

\textbf{LLM Pruning:} In the era of LLMs, methods have been proposed to overcome the challenges mentioned above~\cite{li2023sparse}. Recent research endeavors have evolved towards post-training pruning methods, which start from the pre-trained network and remove redundant parameters \textit{without end-to-end fine-tuning or retraining}. SparseGPT
uses second-order information to address a layer-wise reconstruction problem and prunes large models with unstructured and N:M structured sparsity~\cite{zhou2021learning} respectively. Wanda~\cite{sun2023simple} proposes a new pruning metric that takes both weight magnitude and their corresponding input activations into consideration, achieving comparable perplexity with SparseGPT~\cite{Frantar2023SparseGPTML}. However, ~\cite{jaiswal2023compressing} points out that perplexity is not necessarily an accurate metric for evaluating the effectiveness of model compression, with both SparseGPT and Wanda fail to achieve satisfactory performance even with low-level sparsity (25-30\%). Based on this issue, we speculate that the lack of \textit{retraining} after removing the unimportant weights will lead to an undesirable decline in performance, and put forward a novel training method with high \textit{training efficiency}.

\begin{figure*}[h!]
    \centering
    \vspace{0.3em}
    \includegraphics[width=\textwidth]{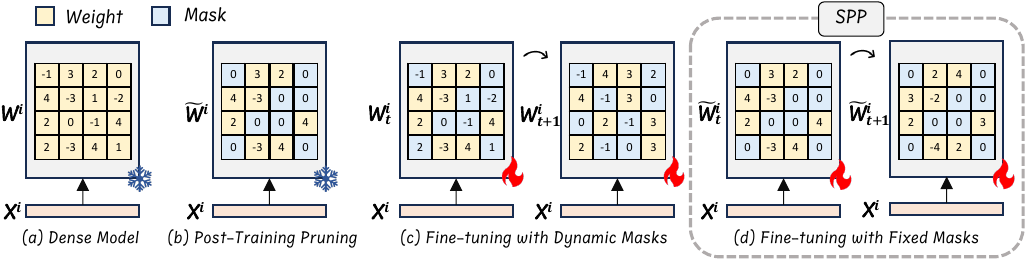}
    \vspace{-1em}
    \caption{Comparison between different methods for sparsification of neural networks. (a) A dense linear layer with matrix $\W^i$ and activation $\X^i$. (b) Post-training pruning methods leverage weight magnitude and calibration data for weight pruning. (c) Model fine-tuning with dynamic weight masks, typically referred to as full fine-tuning methods in the literature, adaptively changes the weight masks during the fine-tuning process. (d) Model fine-tuning with fixed weight masks. Our proposed SPP distinguishes itself as a parameter-efficient fine-tuning algorithm that can consistently maintain model sparsity during both the training and weight-merging phases.} 
    \label{fig:overview}
    % \vspace{-1.2em}
\end{figure*}

\textbf{Parameter-efficient  Fine-tuning:} In different language and vision tasks, the pre-training and fine-tuning paradigms have been proven to be highly effective. Compared with full parameter fine-tuning, Parameter-Efficient Fine-Tuning (PEFT)~\cite{peft,xu2023qa} methods freeze most parameters of \textbf{dense} pre-trained models and aim to exhibit comparable capabilities on downstream tasks.  LoRA~\cite{hu2021lora} introduces trainable low-rank decomposition matrices into dense network weights. Adapters~\cite{houlsby2019parameter} insert
lightweight adaption modules into each block of the language models. Different from
previous efforts for dense pre-trained language models, we propose the SPP method with few learnable parameters, which is specially designed for sparse LLMs.

\section{Method}
\label{method}

The objective of neural network pruning is to preserve the performance of the original network as closely as possible by creating a sparse network through the selective removal of certain neural network parameters~\cite{hassibi1993optimal,Han2015LearningBW}.
In this section, we first introduce the notation used in our paper and categorize some existing algorithms for neural network pruning (especially for LLMs) in Sec.~\ref{sec:method_preliminary}, and then elaborate on our proposed SPP in Sec.~\ref{sec:method_proposed}, which is a parameter-efficient fine-tuning method specifically designed for training sparse LLMs.

% \subsection{Preliminaries}
\subsection{A Revisit of Model Pruning Methods}\label{sec:method_preliminary}

\textbf{Post-Training Pruning:} Starting from a dense pre-trained model (LLMs in our setting), post-training pruning~\cite{Frantar2023SparseGPTML,zhang2023efficient} algorithms remove redundant parameters in a neural network by calculating a set of weight masks $\M=\{\M^0,\M^1,\dots, \M^{N-1}\}$ leveraging weight magnitude $\W=\{\W^0,\W^1,\dots, \W^{N-1}\}$ together with a set of calibration data. As shown in Fig.~\ref{fig:overview} (a), given original dense matrix $\W^i$ $(0\leq i\leq {N-1})$, post-training pruning algorithms compute binary weight masks $\M^i$, and the new set of sparse matrices $\widetilde{\W}$ can be calculated as:\begin{equation}
\resizebox{0.9\hsize}{!}{$\begin{aligned}
\widetilde{\W} = \{ \W^0 \odot \M^0, \W^1 \odot \M^1, ..., \W^{N-1}\odot \M^{N-1} \},
\end{aligned}$}
\end{equation}
where $N$ denotes the number of linear layers, and $\odot$ indicates element-wise multiplication (Hadamard product~\cite{horn1990hadamard}) between matrices. One example based on 2:4 structured sparsity and using weight magnitude as the pruning metric is illustrated in Fig.~\ref{fig:overview} (b). Post-training pruning algorithms are known for their relatively low consumption of computational resources. However, as demonstrated in SparseGPT~\cite{Frantar2023SparseGPTML} and Wanda~\cite{sun2023simple}, even the pruning of LLMs to a moderate level of sparsity, specifically to 50\%, inevitably leads to a substantial degradation in performance~\cite{Han2015LearningBW}. This indicates that after post-training pruning, it is essential to retrain the sparse model to reclaim its knowledge capabilities.
% to the best.

%把post pruning 的模型定义一下 解释一下符号，然后说LLM的压缩即使到middle-level (50\%) 稀疏度也drop很多，cite wanda sparsegpt 举2:4 的例子
\begin{figure*}[h!]
    \centering
    \vspace{0.2em}
    \includegraphics[width=\textwidth]{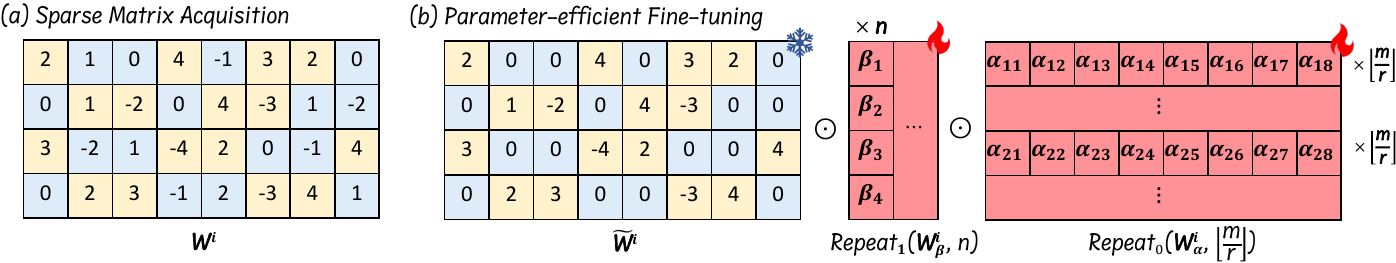}
    \vspace{-0.5em}
    \caption{Learnable parameter insertion of our proposed SPP for a $m\times n$ ($m=4,n=8$) weight matrix. (a) We acquire the mask $\M^i$ of the sparse matrix of a linear layer by pruning with post-training algorithms (e.g. SparseGPT~\cite{Frantar2023SparseGPTML}, Wanda~\cite{sun2023simple}), and obtain sparse linear weights $\widetilde{\W}^i=\W^i\odot \M^i$. (b) SPP adds two sets of parameters to the sparse matrix, one ($\W^i_{\beta}\in \R^{m\times 1}$) initialized to 0 and multiplied on each row of the matrix, and one ($\W^i_{\alpha}\in \R^{r\times n}$, $r=2$ in this example) randomly initialized and multiplied on the columns of the matrix. We then scale $\W^i_{\alpha}$ and $\W^i_{\beta}$ to the size of ($m\times n$) and element-wise multiply then with $\widetilde{\W}^i$.
    }
    \label{fig:peft-detail}
    \vspace{-0.5em}
\end{figure*}
%%%% 这一段我明天和你对一下？？？？ \aj{Different from retraining the pruned model to recover its knowledge}, 
\textbf{Full Fine-tuning with Dynamic Sparse Masks:} Another pruning scheme that starts from a dense model is to full-finetune the original dense model with dynamically updated sparse masks,
as depicted in Fig.~\ref{fig:overview} (c). Typically, to improve the performance of sparse models, researchers introduce various training methods to optimize the sparse weights $\widetilde{\W}$ and the corresponding binary masks $\M$ simultaneously~\cite{lin2020dynamic, bengio2013estimating}:
\begin{align}
\label{eq:ste}
\W^i_{t+1}=\W^i_t-\gamma_t \mathbf{g}(\W^i_t \odot \M^i_t)=\W^i_t-\gamma_t \mathbf{g}(\widetilde{\W}^i_t),
\end{align}
where $\gamma_t$ denotes the learning rate at time step $t$ and function $\mathbf{g}(\cdot)$ calculates the gradients\footnote{We only show the equation of gradient descent for simplicity.}. For small-scale sparse models, such as those based on ResNet and BERT, full fine-tuning methods like STE-based algorithms~\cite{lin2020dynamic,evci2020rigging,zhou2021learning} can maintain high performance at even $\geq75\%$ sparsity. However, these methods usually require substantial computational power and memory resources. This poses a significant challenge in the era of training LLMs. For instance, fully fine-tuning a LLaMA-65B model requires at least 780G of GPU memory~\cite{dettmers2023qlora}, not yet including the additional memory needed to update the sparse masks.% It is not always easy to obtain so many computational resources. % 8 NVIDIA-A100-80G GPU chips will still cast an OOM error.

%%%% 这一段要更加肯定的强调一下说 保持sparsity 然后联系 符号 W and M，下一个section 也是，不能忘记了 W, binary mask M, \widetilde(W)
\textbf{Fine-tuning with Fixed Masks: } As can be learned from the respective drawbacks of the two approaches described above, there is a need to balance model performance restoration and the training overhead. Actually, Eq.~(\ref{eq:ste}) can be decoupled into the iterative optimization of binary masks $\M$ and sparse weights $\widetilde{\W}$. When training with limited computation resources, %sparse weight optimization is considered to be more effective than updating binary masks
optimizing sparse weights with fixed masks is considered effective for knowledge restoration~\cite{liu2018rethinking}. Generally, if we fix $\M_t^i=\M_0^i$ and apply $\widetilde{\W}^i_t={\W}^i_t\odot \M^i_0$ at the $t$-th iteration, Eq.~(\ref{eq:ste}) will be equivalent to the retraining of sparse models with fixed masks: 
\begin{align}
\label{eq:sgd}
\widetilde{\W}^i_{t+1} = \widetilde{\W}^i_t -\gamma_t \mathbf{g}(\widetilde{\W}^i_t)\odot \M^i_0,
\end{align}
but it is still cumbersome to calculate gradients for $\widetilde{\W}^i_t$ associated with fixed mask $\M^i_0$. To this end, we provide the SPP method for efficient fine-tuning post-training pruned LLMs with fixed masks (as illustrated in Fig.~\ref{fig:overview} (d)).

\subsection{Proposed Method}\label{sec:method_proposed}

In this subsection, we propose SPP to balance model performance restoration and computational overhead. Current PEFT methods carry out weight updates of huge linear blocks by introducing trainable low-rank decomposition matrices~\cite{hu2021lora} and adding them with original frozen linear weights after training. This approach finally results in dense matrices and destroys model sparsity. Reflecting on this issue, we suggest that the direct addition of new parameters to sparse matrices leads to the inevitable change in model sparsity, whereas weight {multiplication} can avoid the problem. To this end, we add two sets of lightweight \textit{orthogonal} learnable parameters $\W_{\alpha}=\{\W^0_{\alpha},\W^1_{\alpha},\dots,\W^{N-1}_{\alpha}\}$ and $\W_{\beta}=\{\W^0_{\beta},\W^1_{\beta},\dots,\W^{N-1}_{\beta}\}$, multiply them separately with column and row weights of each linear matrix in the pruned LLM, and perform efficient fine-tuning for sparse matrices with the original pruned weights fixed.

\textbf{Learnable Parameter Insertion:} The core of SPP is how to introduce learnable parameters such that they do not lead to changes in sparsity during the training and weight-merging process. As shown in Fig.~\ref{fig:peft-detail}, given a sparse linear matrix $\widetilde{\W}^i\in \R^{m\times n}$ ($0\leq i\leq N-1$) after pruning, we insert two sets of learnable parameters: $\W^i_{\alpha}\in \R^{r\times n}$ and $\W_{\beta}^i \in \R^{m\times 1}$. $\W^i_{\alpha}$ adds learnability to the weights of each column, while $\W^i_{\beta}$ 
 further expands it to each row\footnote{We set $r>1$ to increase capacity. We discuss setting $\W^i_{\alpha}\in \R^{r\times n}$ instead of $\W^i_{\beta}\in \R^{m\times r}$ in Sec.~\ref{sec:beta} in the Appendix.}. This leads to $(m+rn)$ additional parameters for one sparse matrix. We then scale $\W^i_{\alpha}$ and $\W^i_{\beta}$ to the same size of $\widetilde{\W}^i$ and multiply them with $\widetilde{\W}^i$ in an element-wise manner:
\begin{equation}
\resizebox{0.9\hsize}{!}{$\begin{aligned}
    \widetilde{\mathbbm{\mathbf{W}}}^{i'} = \widetilde{\W}^i\odot \text{Repeat}_0(\W^i_{\alpha}, \left\lfloor \frac{m}{r} \right\rfloor)\odot \text{Repeat}_1(\W^i_{\beta},n),
\end{aligned}$}
\label{eq:method_w_i}
\end{equation}
where $\text{Repeat}_d(\W,k)$ is a function denoting the repetition of each element of $\W$ by $k$ times along the $d$-th dimension, which can be easily implemented by the \verb|torch.repeat_interleave()| function in Python. Besides, there is no need to perform an \textit{explicit} ``repeat'' operation on $\W^i_\beta$, as PyTorch automatically aligns the matrix dimensions when conducting Hadamard product between a matrix and a vector. The floor function $\left\lfloor \cdot \right\rfloor$ indicates integer division. In our implementation, $r$ is set to the numbers that can be divided by $n$ (e.g., choosing $r=4, 8, 16$, etc.). Benefited from element-wise multiplication, the values of 0 in the original sparse matrix will remain unchanged, so $\widetilde{\W}^{i'}$ shares the same sparsity pattern and sparsity ratio with $\widetilde{\W}^i$, one example with a $4\times 8$ matrix and $r=2$ is shown in Fig.~\ref{fig:peft-detail}. In this way, we can train the linear matrices with fixed masks, following the training pattern in Fig.~\ref{fig:overview} (d). It is worth noting that if we set $r=m$, our method performs a full-parameter gradient update on the original sparse matrix.
\begin{figure}[t]
    \centering
    \vspace{0.2em}
    \includegraphics[width=\linewidth]{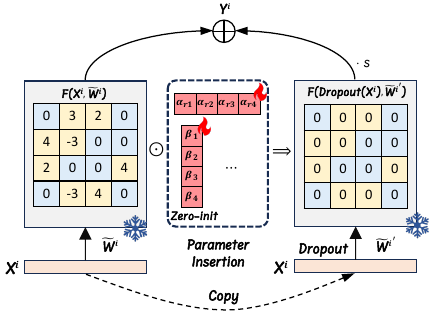}
    \vspace{-0.7em}
    \caption{Our proposed SPP framework for sparse weight models. We multiply two sets of learnable parameters to the frozen linear weights with $\W_\alpha$ randomly initialized and $\W_\beta$ zero-initialized (detailed in Fig.~\ref{fig:peft-detail}). %In this case $\widetilde{\W}^{i'}=\mathbf{0}$ at the time of neural network initialization. 
    Linear operation is conducted on both the original and modified weights, and then the results are added in a residual manner. This framework corresponds to Eq.~(\ref{eq:pipe_sppeft}).}
    \label{fig:peft-pipe}
    \vspace{-0.2em}
\end{figure}
\textbf{Framework of SPP:} After explaining how to insert parameters, we introduce our proposed SPP framework, as shown in Fig.~\ref{fig:peft-pipe}. During the \underline{training process}, in the $i$-th linear layer of the neural network, the output can be calculated by:
\begin{align}
\Y^i=\F(\X^i, \widetilde{\W}^i) + s\cdot \F(\text{Dropout}(\X^i), \widetilde{\W}^{i'}),\label{eq:pipe_sppeft}
\end{align}
where $s$ is a hyper-parameter for scaling, and $\X^i\in \R^{b\times n}$ denotes the input activation at the $i$-th linear layer. We follow Eq.~(\ref{eq:method_w_i}) for the training of $\widetilde{\W}^{i'}$, set initial $\W^i_{\beta}$ to all zeros, and randomly initialize $\W^i_{\alpha}$. In this way, the {\textit{initial structure and weights of the network are consistent with the post-training pruned model}}. More specifically, $\Y^i=\F(\X^i, \widetilde{\W}^i)$ for neural network initialization. This strategy results in more stable training.  %%% 没有强调出来 \Y^i = \F(\X^i, \widetilde{\W}^i) 在刚开始训练的时候
\underline{After training}, the linear layer parameters can be merged as: $\widetilde{\W}^i+s\cdot \widetilde{\W}^{i'}$, sharing the same sparsity pattern and ratio with the original pruned model.
The structure of SPP is similar to LoRA, but it does not destroy the sparse structure of the model during both the training and weight-merging process. For a more straightforward comparison, recall that the formulation of LoRA with frozen weight $\widetilde{\W}^i$ is:
\begin{equation}
\resizebox{0.9\hsize}{!}{$\begin{aligned}
% \begin{align}
    \Y^i=\F(\X^i, \widetilde{\W}^i) + s\cdot \text{Dropout}(\X^i) (\mathbf{A}^i)^T (\mathbf{B}^i)^T,\label{eq:lora}
\end{aligned}$}
\end{equation}
where $\mathbf{A}^i\in \R^{r\times n}$ and $\mathbf{B}^i\in \R^{m\times r}$. During the training and weight-merging process of LoRA, the weights of the linear layer matrix are equivalent to ($\widetilde{\W}^i+s\cdot \mathbf{B}^i\mathbf{A}^i$), which is actually a dense matrix.

\textbf{Memory Usage Optimization:} We find that for each linear layer with input activation $\X^i$, the insertion of $\W^i_{\alpha}$ and $\W^i_{\beta}$ greatly reduces the number of trainable parameters, but we still need to store the intermediate matrix $\text{Repeat}_0(\W^i_{\alpha}, \left\lfloor \frac{m}{r} \right\rfloor)$ when calculating the linear function $\Y^{i'}=\X^i(\widetilde{\W}^{i'})^T$ (this is actually the second addition term in Eq.~(\ref{eq:pipe_sppeft})), which usually takes a huge amount of memory. To overcome this problem, we redesign the procedure of matrix multiplication in SPP following the column parallel linear layer in Megatron-LM~\cite{shoeybi2019megatron}. Specifically, in each linear layer $\F(\X, \W)$ with weight $\W^{m\times n}$ and activation $\X^{b\times n}$, the output of the linear layer is  
$\Y=\X\W^T\in \R^{b\times m}$ (w.l.o.g., we discard the batch of input $\X$ and assume $b=1$ for simplicity). We split the column of $\W$ into $r$ blocks, denoted as $\begin{bmatrix}\W^T_0, \W^T_1, \dots, \W^T_{r-1}\end{bmatrix}^T$, where each $\W_j\in \R^{\left\lfloor \frac{m}{r} \right\rfloor \times n}$. In our setting, we can derive:
\begin{align}
\Y^{i'}&=\X^i(\widetilde{\W}^{i}\odot \text{Repeat}_0(\W^i_{\alpha}, \left\lfloor \frac{m}{r} \right\rfloor)\odot \text{Repeat}_1(\W^i_{\beta},n))^T\nonumber\\
&=\begin{bmatrix}
\cdots (\X^i\odot \W^i_{\alpha j})(\widetilde{\W}^{i}_j)^T\cdots
    % (\X^i\odot \W^i_{\alpha 1})(\widetilde{\W}^{i}_1)^T\cdots(\X^i\odot \W^i_{\alpha r})(\widetilde{\W}^{i}_r)^T
\end{bmatrix}\odot (\W^i_{\beta})^T,
\label{equ:matrix_memory}
\end{align}
where $\W^i_{\alpha j}$ denotes the $j$-th row of $\W^i_{\alpha}$, and $\widetilde{\W}^{i}_j$ denots the $j$-th block of $\widetilde{\W}^i$ ($0\leq j\leq r-1$). This will eliminate the need for saving the matrix $\text{Repeat}_0(\W^i_{\alpha}, \left\lfloor \frac{m}{r} \right\rfloor)$, and a more detailed proof of the above Eq.~(\ref{equ:matrix_memory}) can be found in Sec.~\ref{append:proof_of_matrix_mem} in the Appendix.

\textbf{Remarks on SPP: }

1) Why the residual scheme is needed? Actually, it is simpler to directly initialize $\W_\alpha$ and $\W_\beta$ to 1 and then multiply them to the corresponding positions respectively. But we find that such a training scheme converges more slowly.
%and the final model performance is poor.

2) Why is it important to keep the sparse pattern of the model during training? Previous PEFT schemes tend to turn a sparse model into a dense one after training, and we need to prune the trained model again to obtain the final sparse model. This strategy leads to an unpredictable change in model performance again. In contrast, by keeping the model sparsity during the training process, the performance of the final model is a direct reflection of the effectiveness of the training process~\cite{liu2018rethinking}.

%稍微想一个名字 对应图4 写出 formulation

% 想一下 要不要connection lora 的formulation 加一个区分

% \newpage
\section{Experiments}\label{experiments}
% \usepackage{graphicx}
\begin{table*}[h!]\fontsize{8.5}{9}\selectfont
% \vspace{-0.8em}
\centering
\resizebox{\textwidth}{!}{%
% \begin{tabular}{ccccccccc}
\begin{tabular*}{\hsize}{@{}@{\extracolsep{\fill}}lcccccccc@{}}\midrule
% \midrule
\textbf{} &
  \multicolumn{4}{c}{\textbf{LLaMA}} &
  \multicolumn{3}{c}{\textbf{LLaMA-2}} \\ \midrule
\textbf{} &
  \multicolumn{1}{c}{\text{7B}} &
  \multicolumn{1}{c}{\text{13B}} &
  \multicolumn{1}{c}{\text{30B}} &
  \text{65B} &
  \multicolumn{1}{c}{\text{7B}} &
  \multicolumn{1}{c}{\text{13B}} &
  \text{70B} \\ \midrule
\text{Trainable Parameters} &
  \multicolumn{1}{c}{2.0$\times$10$^7$} &
  \multicolumn{1}{c}{3.1$\times$10$^7$} &
  \multicolumn{1}{c}{6.0$\times$10$^7$} &
  9.8$\times$10$^7$ &
  \multicolumn{1}{c}{2.0$\times$10$^7$} &
  \multicolumn{1}{c}{3.1$\times$10$^7$} &
  1.1$\times$10$^8$ \\ %\midrule
\text{All Parameters} &
  \multicolumn{1}{c}{6.8$\times$10$^9$} &
  \multicolumn{1}{c}{1.3$\times$10$^{10}$} &
  \multicolumn{1}{c}{3.3$\times$10$^{10}$} &
  6.5$\times$10$^{10}$ &
  \multicolumn{1}{c}{6.8$\times$10$^9$} &
  \multicolumn{1}{c}{1.3$\times$10$^{10}$} &
  6.9$\times$10$^{10}$ \\ %\midrule
\textbf{Per mille (\textperthousand)} &
  \multicolumn{1}{c}{\textbf{2.90}} &
  \multicolumn{1}{c}{\textbf{2.35}} &
  \multicolumn{1}{c}{\textbf{1.83}} &
  \textbf{1.50} &
  \multicolumn{1}{c}{\textbf{2.90}} &
  \multicolumn{1}{c}{\textbf{2.35}} &
  \textbf{1.54} \\ \midrule
\end{tabular*}%
}
\vspace{-0.5em}
\caption{Number of trainable parameters, total number of parameters, and per mille of trainable parameters of SPP with $r=16$. We only need to fine-tune a small fraction of parameters compared with full fine-tuning.}
\label{tab:params_number}
% \vspace{-0.5em}
\end{table*}

% \usepackage{multirow}
% \usepackage{graphicx}
\begin{table*}[h!]
\centering
\resizebox{\textwidth}{!}{%
\begin{tabular}{clccccccccc}
\midrule
\textbf{LLaMA} &
  \textbf{Method} &
  \textbf{Sparsity} &
  \textbf{BoolQ} &
  \textbf{RTE} &
  \textbf{HellaSwag} &
  \textbf{WinoGrande} &
  \textbf{ARC-e} &
  \textbf{ARC-c} &
  \textbf{OBQA} &
  \textbf{Average} \\ \midrule
\multirow{9}{*}{\textbf{7B}}  & None            & Dense            & 75.11 & 66.43 & 56.96 & 69.85 & 75.25 & 41.89 & 34.40 & 59.98 \\ \cmidrule{2-11} 
                              & SparseGPT      & Unstructured 50\% & \gc{73.36} & \gc{57.76} & \gc{51.44} & \gc{68.03} & \gc{70.45} & \gc{36.35} & \gc{28.40} & \gc{55.11} \\ %\cmidrule{2-11} 
                              & SparseGPT\textbf{+SPP} & Unstructured 50\% & 72.84 & 65.70 & 56.40 & 67.88 & 72.35 & 41.04 & 32.80 & \textbf{58.43} \\ %\cmidrule{2-11} 
                              & SparseGPT      & 2:4              & \gc{70.09} & \gc{57.76} & \gc{43.37} & \gc{63.46} & \gc{61.62} & \gc{29.27} & \gc{22.60} & \gc{49.74} \\ %\cmidrule{2-11} 
                              & SparseGPT\textbf{+SPP} & 2:4              & 72.39 & 59.57 & 53.33 & 64.17 & 68.39 & 37.54 & 26.80 & \textbf{54.60} \\ \cmidrule{2-11} 
                              & Wanda          & Unstructured 50\% & \gc{71.01} & \gc{55.23} & \gc{51.90} & \gc{66.22} & \gc{69.36} & \gc{36.95} & \gc{28.60} & \gc{54.18} \\ %\cmidrule{2-11} 
                              & Wanda\textbf{+SPP}     & Unstructured 50\% & 70.86 & 66.06 & 55.92 & 67.64 & 72.81 & 41.64 & 32.00 & \textbf{58.13} \\ %\cmidrule{2-11} 
                              & Wanda          & 2:4              & \gc{69.27} & \gc{51.26} & \gc{42.07} & \gc{62.67} & \gc{60.52} & \gc{27.99} & \gc{24.60} & \gc{48.34} \\ %\cmidrule{2-11} 
                              & Wanda\textbf{+SPP}     & 2:4              & 71.19 & 63.90 & 52.77 & 64.88 & 68.18 & 37.03 & 30.00 & \textbf{55.42} \\ \midrule
\multirow{9}{*}{\textbf{13B}} & None            & Dense            & 77.98 & 70.40 & 59.92 & 72.61 & 77.36 & 46.50 & 33.20 & 62.57 \\ \cmidrule{2-11} 
                              & SparseGPT      & Unstructured 50\% & \gc{76.54} & \gc{62.09} & \gc{54.94} & \gc{71.59} & \gc{72.35} & \gc{41.64} & \gc{32.20} & \gc{58.76} \\ %\cmidrule{2-11} 
                              & SparseGPT\textbf{+SPP} & Unstructured 50\% & 79.20 & 64.62 & 59.27 & 70.32 & 74.83 & 46.59 & 34.60 & \textbf{61.35} \\ %\cmidrule{2-11} 
                              & SparseGPT      & 2:4              & \gc{70.80} & \gc{56.68} & \gc{48.09} & \gc{69.22} & \gc{66.88} & \gc{36.26} & \gc{26.20} & \gc{53.45} \\ %\cmidrule{2-11} 
                              & SparseGPT\textbf{+SPP} & 2:4              & 77.65 & 63.54 & 56.55 & 69.69 & 71.21 & 40.96 & 32.60 & \textbf{58.89} \\ \cmidrule{2-11} 
                              & Wanda          & Unstructured 50\% & \gc{76.27} & \gc{62.82} & \gc{55.78} & \gc{71.98} & \gc{73.32} & \gc{43.77} & \gc{31.80} & \gc{59.39} \\ %\cmidrule{2-11} 
                              & Wanda\textbf{+SPP}     & Unstructured 50\% & 78.29 & 66.43 & 58.88 & 70.32 & 75.59 & 46.93 & 34.40 & \textbf{61.55} \\ %\cmidrule{2-11} 
                              & Wanda          & 2:4              & \gc{70.21} & \gc{53.79} & \gc{46.78} & \gc{68.82} & \gc{65.74} & \gc{33.70} & \gc{26.20} & 52.18 \\ %\cmidrule{2-11} 
                              & Wanda\textbf{+SPP}     & 2:4              & 75.99 & 58.12 & 56.07 & 68.90 & 70.37 & 40.53 & 32.40 & \textbf{57.48} \\ \midrule
\multirow{9}{*}{\textbf{30B}} & None            & Dense            & 82.63 & 66.79 & 63.36 & 75.85 & 80.39 & 52.82 & 36.00 & 65.41 \\ \cmidrule{2-11} 
                              & SparseGPT      & Unstructured 50\% & \gc{82.63} & \gc{58.84} & \gc{59.20} & \gc{73.48} & \gc{78.79} & \gc{49.15} & \gc{33.20} & \gc{62.18} \\ %\cmidrule{2-11} 
                              & SparseGPT\textbf{+SPP} & Unstructured 50\% & 84.43 & 68.23 & 63.18 & 73.56 & 81.57 & 52.56 & 37.00 & \textbf{65.79} \\ %\cmidrule{2-11} 
                              & SparseGPT      & 2:4              & \gc{76.57} & \gc{61.01} & \gc{53.52} & \gc{72.30} & \gc{74.66} & \gc{42.06} & \gc{31.60} & \gc{58.82} \\ %\cmidrule{2-11} 
                              & SparseGPT\textbf{+SPP} & 2:4              & 81.65 & 66.43 & 60.46 & 72.45 & 78.75 & 50.17 & 36.20 & \textbf{63.73} \\ \cmidrule{2-11} 
                              & Wanda          & Unstructured 50\% & \gc{81.93} & \gc{64.98} & \gc{60.95} & \gc{73.64} & \gc{79.38} & \gc{50.17} & \gc{34.80} & \gc{63.69} \\ %\cmidrule{2-11} 
                              & Wanda\textbf{+SPP}     & Unstructured 50\% & 84.19 & 66.79 & 62.52 & 71.59 & 77.10 & 51.79 & 34.80 & \textbf{64.11} \\ %\cmidrule{2-11} 
                              & Wanda          & 2:4              & \gc{75.14} & \gc{63.54} & \gc{54.53} & \gc{72.45} & \gc{74.24} & \gc{41.89} & \gc{31.80} & \gc{59.08} \\ %\cmidrule{2-11} 
                              & Wanda\textbf{+SPP}     & 2:4              & 81.38 & 69.68 & 59.99 & 71.59 & 76.73 & 48.63 & 34.60 & \textbf{63.23} \\ \midrule
\multirow{9}{*}{\textbf{65B}} & None            & Dense            & 84.55 & 69.68 & 65.40 & 77.35 & 52.82 & 81.00 & 38.00 & 66.97 \\ \cmidrule{2-11} 
                              & SparseGPT      & Unstructured 50\% & \gc{84.90} & \gc{70.04} & \gc{63.95} & \gc{77.27} & \gc{79.65} & \gc{50.17} & \gc{37.40} & \gc{66.20} \\ %\cmidrule{2-11} 
                              & SparseGPT\textbf{+SPP} & Unstructured 50\% & 84.95 & 70.04 & 64.25 & 77.19 & 79.85 & 50.94 & 37.80 & \textbf{66.43} \\ %\cmidrule{2-11} 
                              & SparseGPT      & 2:4              & \gc{84.55} & \gc{69.31} & \gc{57.95} & \gc{76.95} & \gc{78.00} & \gc{45.39} & \gc{31.20} & \gc{63.34} \\ %\cmidrule{2-11} 
                              & SparseGPT\textbf{+SPP} & 2:4              & 84.25 & 68.23 & 58.40 & 76.87 & 78.10 & 45.99 & 31.40 & 63.32 \\ \cmidrule{2-11} 
                              & Wanda          & Unstructured 50\% & \gc{85.05} & \gc{71.84} & \gc{64.60} & \gc{77.35} & \gc{79.65} & \gc{50.26} & \gc{38.40} & \gc{66.74} \\ %\cmidrule{2-11} 
                              & Wanda\textbf{+SPP}     & Unstructured 50\% & 85.25 & 71.84 & 65.30 & 77.19 & 79.95 & 51.11 & 38.60 & \textbf{67.03} \\ %\cmidrule{2-11} 
                              & Wanda          & 2:4              & \gc{83.40} & \gc{61.01} & \gc{58.55} & \gc{75.22} & \gc{76.60} & \gc{45.56} & \gc{33.20} &\gc{61.93} \\ %\cmidrule{2-11} 
                              & Wanda\textbf{+SPP}     & 2:4              & 83.30 & 61.37 & 61.85 & 76.16 & 78.60 & 47.70 & 36.20 & \textbf{63.60} \\ \midrule
\end{tabular}%
}
\vspace{-0.5em}
\caption{Zero-shot evaluation results of 7 tasks from EleutherAI LM Harness~\cite{gao2021framework} after training LLaMA on Alpaca~\cite{taori2023alpaca} dataset by SPP. SPP improves the performance of sparse models from SparseGPT and Wanda. In 2 cases, the performance of sparsely trained models even exceeds their dense counterparts.}
\label{tab:zero_llama1_50}
\vspace{-1em}
\end{table*}

In this section, we report a series of experiments to demonstrate the effectiveness of SPP for the efficient training of sparse pruned models. 

\textbf{Experiment Setup: } We choose LLaMA and LLaMA-2 model families: LLaMA 7B/13B/30B/65B~\cite{touvron2023llama1} and LLaMA-2 7B/13B/70B~\cite{touvron2023llama2} for experiments. For these LLMs, we first prune them using post-training pruning algorithms SparseGPT~\cite{Frantar2023SparseGPTML} and Wanda~\cite{sun2023simple}, then carry out instruction fine-tuning for the recovery of knowledge. All the training and testing processes are conducted on a server with 8 NVIDIA A100-80GB GPUs\footnote{This is the largest computing resource we have access to. Fine-tuning schemes outlined in Eq.~(\ref{eq:ste}) and (\ref{eq:sgd}) result in Out-of-Memory errors when applied to models with more than 30B parameters.}.

\textbf{Training and Evaluation Details: } We use high quality instruction fine-tuning dataset Stanford-Alpaca~\cite{taori2023alpaca} to train the pruned models.%(for training LLaMA-2 models) and Alpaca-GPT4~\cite{peng2023instruction} (for training LLaMA models). 
We do not use pre-training datasets (e.g. C4~\cite{2020t5}, SlimPajama~\cite{cerebras2023slimpajama}, etc.) as they are quite large but of low quality. Unlike previous related works on training sparse models (e.g. Sheared llama~\cite{xia2023sheared}), our approach requires only several hours of training on a smaller dataset. 
We add learnable parameters on ``q\_proj, k\_proj, v\_proj, o\_proj, gate\_proj, up\_proj, down\_proj, score'' linear layers, and $r$ is set to 16 %
% in all our experiments 
(unless specifically noted). For training convenience, we fine-tune 7B/13B/30B models by 3 epochs, and 65B/70B models by 1 epoch. In terms of evaluation, we mainly report zero-shot performance on seven tasks from EleutherAI LM Harness~\cite{gao2021framework} following~\cite{sun2023simple}. We also test few-shot performances on the MMLU~\cite{hendrycks2021ethics,hendryckstest2021} benchmark.

% \usepackage{multirow}
% \usepackage{graphicx}
\begin{table}[t]\huge
\centering
\vspace{-0.1em}
\resizebox{\columnwidth}{!}{%
\begin{tabular}{clccc}
\midrule
{ \textbf{LLaMA}} &
  { \textbf{Method}} &
  { \textbf{Sparsity}} &
  { \textbf{LM-eval}} & {\textbf{PPL ($\mathbf{\downarrow}$)}} \\ \midrule
 &
   { Wanda} &
  { Unstructured 75\%} &
  32.14& {1285.24} \\ % \cmidrule{2-4} 
&
  { Wanda+DS$\oslash$T} &
  { Unstructured 75\%} &
  32.23 & { 646.40} \\ % \cmidrule{2-4} 
 &
  { Wanda\textbf{+SPP}} &
  { Unstructured 75\%} &
  \textbf{41.71}&\textbf{ 21.80} \\ \cmidrule{2-5}
 &
    { Wanda} &
  { 2:8} &
  32.53& { 3284.43} \\ % \cmidrule{2-4} 
&
  { Wanda+DS$\oslash$T} &
  { 2:8} &
  31.57 & { 2742.98} \\  
\multirow{-6}{*}{{ \textbf{7B}}} &
  { Wanda\textbf{+SPP}} &
  { 2:8} &
  \textbf{38.61}& \textbf{ 42.07 } \\ \midrule
 &
   { Wanda} &
  { Unstructured 75\%} &
  39.21& { 149.63} \\
  &
  { Wanda+DS$\oslash$T} &
  { Unstructured 75\%} &
  37.77 & { 184.51} \\ % \cmidrule{2-4} 
 &
  { Wanda\textbf{+SPP}} &
  { Unstructured 75\%} &
  \textbf{50.33}& \textbf{ 10.89} \\ \cmidrule{2-5} 
 &
   { Wanda} &
  { 2:8} &
  35.12& { 1057.58} \\ &
  { Wanda+DS$\oslash$T} &
  { 2:8} &
32.81 & 903.17 \\ % \cmidrule{2-4} 
\multirow{-6}{*}{{ \textbf{30B}}} &
  { Wanda\textbf{+SPP}} &
  { 2:8} &
  \textbf{43.09} & \textbf{19.83} \\ \bottomrule
\end{tabular}%
}
\vspace{-0.3em}
\caption{Comparison for pruning LLaMA model family at 75\% sparsity. We provide average accuracies of 7 zero-shot tasks together with WikiText perplexity~\cite{merity2016pointer}. SPP achieves far better results than the other two post-training pruning methods.}
\label{tab:compare_dsnot}
\vspace{-1.2em}
\end{table}

\textbf{Model Sparsity and Baselines: } For most experiments, we use the 50\% sparsity ratio, including unstructured 50\% sparsity and 2:4 sparsity~\cite{mishra2021accelerating,zhou2021learning}. We compare the models trained by SPP with post-training pruned models (SparseGPT~\cite{Frantar2023SparseGPTML}, Wanda~\cite{sun2023simple}) and the original dense models. To extend the experiments to higher sparsity, we test some of the 75\% sparsity cases, including unstructured 75\% sparsity and 2:8 sparsity, and compare our SPP with LoRA* and recently proposed DS$\oslash$T~\cite{zhang2023dynamic}. %by sparsifying the model after LoRA training.

\subsection{Number of Trainable Parameters} 

An important metric for estimating a PEFT method is the number of learnable parameters during training. Tab.~\ref{tab:params_number} summarizes the number and per mille (\textperthousand) of trainable parameters in our experiments with $r=16$. As can be seen,  SPP requires training only a very small fraction of parameters. % Moreover, since we fix $r=16$ in all the above experiments, the percentage of learnable parameters decreases as the number of parameters increases in the original dense model (except for LLaMA2-70B, which uses a different structure than the other LLaMA models).

\subsection{Zero-shot Evaluation Results}\label{sec:zero_shot}

In Tab.~\ref{tab:zero_llama1_50} and Tab.~\ref{tab:zero_llama2_50} (Tab.~\ref{tab:zero_llama2_50} is in Sec.~\ref{sec:llama2_res} in the Appendix due to space limitation), we show the zero-shot accuracies of post-training pruned models and their retrained counterparts by SPP on both LLaMA and LLaMA-2 (task-wise and average results). We follow the evaluation tasks and experiment settings of Wanda~\cite{sun2023simple}. As seen from the tables, after SPP training, the average performances of the vast majority of models are improved. Especially for the models with 2:4 sparsity, we observe around 8\% average performance improvement, larger than that for the unstructured 50\% sparsity models. By utilizing the cuSPARSELt %~\footnote{\url{https://docs.nvidia.com/cuda/cusparselt/}} 
library, the N:M technique can be efficiently implemented on NVIDIA Ampere Graphics processors, resulting in practical speedups. The enhancement of 2:4 sparsity models is thus of particular importance.
% \usepackage{multirow}
% \usepackage{graphicx}
\begin{table}[t]\huge
\centering
\vspace{-0.1em}
\resizebox{\columnwidth}{!}{%
\begin{tabular}{cccccc}
\midrule
\textbf{Method} & \textbf{Sparsity}& \textbf{Zero-init} & \textbf{$\W_\beta$} & \textbf{$r$} & \textbf{LM-eval} \\ \midrule
\multirow{10}{*}{{\makecell{Wanda\textbf{+SPP}}}} & \multirow{5}{*}{2:4} & \checkmark & \checkmark & 4 & 54.04 \\ %\cmidrule{3-6} 
 & & \checkmark & \checkmark & 8 & 54.87 \\ %\cmidrule{3-6} 
 & & \checkmark & & 16 & 54.52  \\ %\cmidrule{3-6} 
 & & &\checkmark & 16 & 53.52 \\ % \cmidrule{2-6} 
 & & \checkmark & \checkmark & 16 & \textbf{55.42} \\ \cmidrule{2-6} 
 & \multirow{5}{*}{\makecell{Unstructured 50\%}} & \checkmark &\checkmark & 4 & 57.86 \\ %\cmidrule{3-6} 
 & & \checkmark & \checkmark & 8 & 56.39 \\ %\cmidrule{3-6} 
 & & \checkmark & & 16 & 57.81\\ %\cmidrule{3-6} 
 & & &\checkmark & 16 & 57.59 \\ % \cmidrule{2-6} 
 & & \checkmark & \checkmark & 16 & \textbf{58.13}  \\ \midrule
\multirow{10}{*}{{\makecell{SparseGPT\textbf{+SPP}}}} & \multirow{5}{*}{2:4} & \checkmark & \checkmark& 4 & \textbf{54.82} \\ %\cmidrule{3-6} 
 & & \checkmark & \checkmark & 8 & 54.24\\ %\cmidrule{3-6} 
 & & \checkmark & & 16 & {54.62}\\ %\cmidrule{3-6} 
 & & &\checkmark & 16 & 54.01 \\ % \cmidrule{2-6} 
 & & \checkmark & \checkmark & 16 & 54.60\\ \cmidrule{2-6} 
 & \multirow{5}{*}{{\makecell{Unstructured 50\%}}} & \checkmark & \checkmark & 4 & 57.58 \\ %\cmidrule{3-6} 
 & & \checkmark & \checkmark & 8 & 57.32 \\ %\cmidrule{3-6} 
 & & \checkmark & & 16 & 57.66 \\ %\cmidrule{3-6} 
 & & &\checkmark & 16 & 57.12 \\ % \cmidrule{2-6} 
 & & \checkmark & \checkmark & 16 & \textbf{58.43}  \\ \midrule
\end{tabular}%
}
\vspace{-0.2em}
\caption{Ablation studies on LLaMA-7B. We investigate the utility of zero-initialization of weights, evaluate the influence of parameter $\W_\beta$, as well as test the performance of SPP-trained models for different values of $r$. Training with zero-init weights, $\W_\beta$, and using $r=16$ get the overall best results in our experiments.}
\label{tab:ablation_study}
\vspace{-0.8em}
\end{table}

\subsection{Extend to Higher Sparsity}

To further validate the effectiveness of our SPP method, we extend the existing post-training pruning method Wanda~\cite{sun2023simple} to 75\% sparsity ratio with unstructured 75\% and 2:8 sparsity. The results are shown in Tab.~\ref{tab:compare_dsnot}. ``LM-eval'' stands for zero-shot evaluation results of 7 different tasks from EleutherAI LM Harness~\cite{gao2021framework}, ``PPL'' stands for Wikitext perplexity~\cite{merity2016pointer}. We find that Wanda~\cite{sun2023simple} with unstructured 75\% and 2:8 sparsity has a significant performance drop, while the dynamic mask without retraining method DS$\oslash$T~\cite{zhang2023dynamic} has limited performance improvement. We then apply the SPP method to the pruned models by Wanda and obtain far better results. This further highlights the effectiveness of parameter retraining -- just as SPP does -- in boosting the performance of sparse LLMs.
% \usepackage{multirow}
% \usepackage{graphicx}
\begin{table*}[t]
\centering
\resizebox{\textwidth}{!}{%
\begin{tabular}{ccccccccccc}
\midrule
\textbf{LLaMA} &
  \textbf{Method} &
  \textbf{Sparsity} &
  \textbf{BoolQ} &
  \textbf{RTE} &
  \textbf{HellaSwag} &
  \textbf{WinoGrande} &
  \textbf{ARC-e} &
  \textbf{ARC-c} &
  \textbf{OBQA} &
  \textbf{Average} \\ \midrule
\multirow{4.5}{*}{\textbf{7B}}
                              & Wanda+LoRA*  & Unstructured 75\%  & 62.39 & 53.07 & 31.81 & 52.64 & 38.22 & 21.08 & 14.60 & 39.11 \\
                              & Wanda\textbf{+SPP} & Unstructured 75\%   & 60.67 & 56.32 & 35.06 & 52.64 & 47.05 & 22.44 & 17.80 & \textbf{41.71} \\ 
 \cmidrule{2-11} 
                               & Wanda+LoRA*  & 2:8               & 61.47 & 53.07 & 29.18 & 53.12 & 34.68 & 20.22 & 14.60 & 38.05 \\
                              & Wanda\textbf{+SPP} & 2:8         & 54.50 & 59.21 & 31.29 & 52.09 & 37.46 & 19.11 & 16.60 & \textbf{38.61} \\ 
 \midrule
\multirow{4.5}{*}{\textbf{30B}}
                              & Wanda+LoRA*  & Unstructured 75\%   & 65.08 & 55.60 & 44.35 & 62.27 & 60.69 & 29.18 & 23.00 & 48.60 \\
                              & Wanda\textbf{+SPP} & Unstructured 75\%   & 67.95 & 54.15 & 47.28 & 62.51 & 63.68 & 30.72 & 26.00 & \textbf{50.33} \\  \cmidrule{2-11} 
                              & Wanda+LoRA*  & 2:8                & 62.17 & 52.71 & 35.96 & 54.30 & 48.48 & 23.21 & 16.20 & 41.86 \\
                              & Wanda\textbf{+SPP} & 2:8                & 62.05 & 54.87 & 38.17 & 55.09 & 49.62 & 23.63 & 18.20 & \textbf{43.09} \\  \midrule
\end{tabular}%
}
% \vspace{-0.5em}
\caption{Zero-shot evaluation of 7 different tasks from EleutherAI LM Harness~\cite{gao2021framework} with 75\% sparsity. We compare SPP with sparse models after LoRA training by applying the dense trained model with the original Wanda pruning masks. SPP achieves overall higher zero-shot performances than LoRA. (*) denotes sparse models obtained from applying Wanda pruning masks after LoRA training.}
\label{tab:high_sparsity}
% \vspace{-0.5em}
\end{table*}

% \usepackage{graphicx}
\begin{table*}[t]\fontsize{8.5}{9}\selectfont
\centering
\resizebox{\textwidth}{!}{%
% \begin{tabular}{ccccccccc}
\begin{tabular*}{\hsize}{@{}@{\extracolsep{\fill}}lcccccccc@{}}\midrule
% \begin{tabular}{ccccccccc}
               &                  & \multicolumn{4}{c}{\textbf{LLaMA}} & \multicolumn{3}{c}{\textbf{LLaMA-2}}       \\\midrule
\textbf{Method}      & \textbf{Sparsity}         & 7B    & 13B   & 30B   & 65B   & {7B}        & {13B}   & {70B}   \\\midrule
None            & Dense            & 35.64 & 47.63 & 58.58 & 63.78 & 46.56     & 55.30  & 69.56 \\\midrule
SparseGPT      & Unstructured 50\% & \gc{32.19} & \gc{40.44} & \gc{52.62} & \gc{59.37} & \gc{36.41}     & \gc{47.47} & \gc{65.57} \\
SparseGPT\textbf{+SPP} & Unstructured 50\% & 30.77 & 43.91 & 54.73 & 59.38 & 39.78     & 48.31 & 65.60  \\\midrule
SparseGPT      & 2:4            & \gc{28.24} & \gc{32.31} & \gc{43.79} & \gc{49.79} & \gc{29.16}     & \gc{38.41} & \gc{57.66} \\
SparseGPT\textbf{+SPP} & 2:4            & 27.81 & 37.55 & 49.01 & 49.50  & 33.28     & 45.63 & 57.85 \\\midrule
Wanda          & Unstructured 50\% & \gc{31.50}  & \gc{39.43} & \gc{52.84} & \gc{58.75} & \gc{34.20}      & \gc{47.78} & \gc{64.45} \\
Wanda\textbf{+SPP}     & Unstructured 50\% & 31.74 & 43.34 & 53.89 & 59.02 & 38.08     & 48.97 & 64.39 \\\midrule
Wanda          & 2:4            & \gc{27.14} & \gc{31.26} & \gc{41.36} & \gc{45.68} & \gc{28.33}     & \gc{35.16} & \gc{56.86} \\
Wanda\textbf{+SPP}     & 2:4            & 28.56 & 35.73 & 46.19 & 47.67 & 30.47     & 42.79 & 57.98\\\midrule
\end{tabular*}%
}
% \vspace{-1em}
\caption{5-shot evaluation of MMLU~\cite{hendrycks2021ethics,hendryckstest2021}, with LLaMA and LLaMA-2 model families trained on Alpaca~\cite{taori2023alpaca} dataset. The performances of sparse models are improved after instruction fine-tuning.}
\label{tab:few_shot_llama1_llama2}
% \vspace{-0.5em}
\end{table*}

\subsection{Ablation Study}

In this subsection, we carry out an ablation study on the LLaMA-7B model. We train the sparse models obtained from Wanda~\cite{sun2023simple} and SparseGPT~\cite{Frantar2023SparseGPTML} at unstructured 50\% and 2:4 sparsity, respectively. We test the performance of the models with and without the $\W_\beta$ parameter, as well as the average zero-shot accuracies of the models for different values of $r$. We also evaluate the effectiveness of the zero-initialization of added weights. The results are shown in Tab.~\ref{tab:ablation_study}. Although different parameter settings may lead to different results, training with $\W_\beta$ and using $r=16$ can obtain the overall best results in our experiments. In the real application of SPP, appropriate parameters can be selected according to the tasks and available computational resources.

\subsection{More Analyses}

In this part, more experiments and analyses are provided.

\textbf{Comparison with LoRA*: }%During our experiments, in addition to training and testing at 50\% sparsity, we also investigate the 75\% sparsity case. The test results of the 7B and 30B LLaMA models at unstructured 75\% and structured 2:8 sparsity cases are shown in Tab.~\ref{tab:high_sparsity}. We report the zero-shot accuracies on 7 tasks, as well as Wikitext PPL~\footnote{A sparsity ratio at 50\% will not lead to dumb PPL results, as opposed to the several hundred to thousand-fold results obtained through the post-training pruning methods at 75\% sparsity level.}~\cite{merity2016pointer} before and after training to show that parameter fine-tuning at high sparsity levels will result in large performance gains. It can be seen that all models get different degrees of performance improvement. With the 75\% sparsity experiments, we want to claim that parameter fine-tuning will result in greater performance gains than performing single mask updating, which will be discussed in detail in Sec.~\ref{conclusion}. In this part, 
We also compare SPP with LoRA*. We adopt $r=8$ in LoRA and add adapters to the same places as SPP, leading to similar numbers of learnable parameters (around 0.28\% for 7B models and 0.18\% for 30B models). Since LoRA yields a dense model after training, to maintain the model sparsity, we apply the original Wanda pruning masks to sparse it (the LoRA training-then-pruning method is denoted as LoRA*) and test at 75\% sparsity. As shown in Tab.~\ref{tab:high_sparsity}, compared to the model pruned after LoRA training, SPP has performance leads in most cases. This highlights the significance of maintaining weight sparsity during the training and weight-merging processes. 

\textbf{Few-shot Results:} 5-shot average results of different models on the MMLU~\cite{hendrycks2021ethics,hendryckstest2021} benchmark are shown in Tab.~\ref{tab:few_shot_llama1_llama2}. Compared to the tasks in Sec.~\ref{sec:zero_shot}, MMLU is a more difficult benchmark and contains some more difficult questions (e.g., math questions). As can be seen from the table, for the post-training pruned models, SPP also brings large performance gains in the vast majority of cases. However, the sparse models trained by SPP still have considerable gaps in performance compared to the original dense models. It should be noted that we obtain performance improvements by instruction fine-tuning on a small-scale dataset. These models would further improve their performance if we extend the scale of the training dataset~\cite{cerebras2023slimpajama}, but it is currently outside the scope of our considerations in this paper.

\textbf{Inference Speedup: }The inference speedup for sparse models is only dependent on the sparsity pattern of the pruned model. Since SPP does not change the sparsity pattern and ratio of the pruned model during training and merging of parameters, it will lead to the same inference speedup as the original pruned model. We refer the readers to Section 4.3 of Wanda~\cite{sun2023simple} paper for results on the relevant inference speedups, where around 1.6$\times$ speed up is noted for linear layers in LLMs and 1.24$\times$ end-to-end latency speedup is observed for LLaMA-7B (\textit{from 312ms to 251ms}) with the structured 2:4 sparsity.

\textbf{Hyper Parameters: }For training 7B/13B/30B/65B/70B models, we use learning rates of 4e-3/2e-3/4e-3/5e-4/5e-4 with per-device batch size set to 8/4/16/8/8. Following~\cite{dettmers2023qlora}, we set a 0.03 warm-up ratio, but decay the learning rate after reaching the targeted peak value. We use the AdamW optimizer with default setting in the Transformers package\footnote{https://github.com/huggingface/transformers} and add a 0.001 weight decay.

After fixing all the hyper-parameters, we train post-training pruned LLMs obtained by SparseGPT~\cite{Frantar2023SparseGPTML} and Wanda~\cite{sun2023simple} with different sparsity patterns and ratios. We do not carry out hyper-parameter tuning for specific sparsity patterns or ratios. Experiment results shown in the paper demonstrate that our SPP can bring about stable performance improvements.

% \newpage
\section{Conclusion and Discussion}
\label{conclusion}

In this paper, we introduce the novel \textbf{S}parsity-\textbf{P}reserved \textbf{P}arameter-efficient fine-tuning (SPP) method, which is a notably efficient approach for retraining or fine-tuning sparse models to tackle the challenge of restoring the performance of LLMs after pruning. 

Before the advent of LLMs, the field of model pruning primarily explored methods to identify optimal binary sparse masks while training the remaining parameters, either through gradient-based techniques or heuristic approaches. Our research incorporates the SPP method into existing post-training pruning strategies that utilize fixed masks, and focuses solely on retraining the preserved parameters after pruning. Looking ahead, we aim to further develop our SPP method, and integrate it with iterative mask updating techniques to enhance the performance of sparsity-preserved retraining. We hope this strategy can further boost the development in this field of study.

\section*{Impact Statement}
This paper presents work whose goal is to advance the field of Machine Learning, especially the field of large language models (LLMs). Our work can contribute to reducing the computational resources and energy consumption required for operating these huge models, thereby addressing environmental concerns associated with large-scale deep network operations. Additionally, the improved efficiency of LLMs can facilitate more widespread and accessible applications of AI technologies, potentially democratizing the benefits of AI across various sectors. However, as with any advancement in machine learning, there is a need for ongoing consideration of ethical implications, particularly in terms of data privacy, security, and the potential for unintended biases in model outputs. These are questions that all of us in the Machine Learning community need to consider when moving forward.

\section*{Acknowledgement}
This project is funded in part by National Key R\&D Program of China Project 2022ZD0161100, by the Centre for Perceptual and Interactive Intelligence (CPII) Ltd under the Innovation and Technology Commission (ITC)’s InnoHK, by General Research Fund of Hong Kong RGC Project 14204021. Hongsheng Li is a PI of CPII under the InnoHK. 

\nocite{langley00}

\bibliography{example_paper}
\bibliographystyle{icml2024}

%%%%%%%%%%%%%%%%%%%%%%%%%%%%%%%%%%%%%%%%%%%%%%%%%%%%%%%%%%%%%%%%%%%%%%%%%%%%%%%
%%%%%%%%%%%%%%%%%%%%%%%%%%%%%%%%%%%%%%%%%%%%%%%%%%%%%%%%%%%%%%%%%%%%%%%%%%%%%%%
% APPENDIX
%%%%%%%%%%%%%%%%%%%%%%%%%%%%%%%%%%%%%%%%%%%%%%%%%%%%%%%%%%%%%%%%%%%%%%%%%%%%%%%
%%%%%%%%%%%%%%%%%%%%%%%%%%%%%%%%%%%%%%%%%%%%%%%%%%%%%%%%%%%%%%%%%%%%%%%%%%%%%%%

\newpage
\appendix
\onecolumn

\section{Appendix}\label{appendix}

% You can have as much text here as you want. The main body must be at most $8$ pages long.
% For the final version, one more page can be added.
% If you want, you can use an appendix like this one.  

% The $\mathtt{\backslash onecolumn}$ command above can be kept in place if you prefer a one-column appendix, or can be removed if you prefer a two-column appendix.  Apart from this possible change, the style (font size, spacing, margins, page numbering, etc.) should be kept the same as the main body.
%%%%%%%%%%%%%%%%%%%%%%%%%%%%%%%%%%%%%%%%%%%%%%%%%%%%%%%%%%%%%%%%%%%%%%%%%%%%%%%
%%%%%%%%%%%%%%%%%%%%%%%%%%%%%%%%%%%%%%%%%%%%%%%%%%%%%%%%%%%%%%%%%%%%%%%%%%%%%%%

\subsection{More explanations}\label{sec:beta}

Discussion of setting $\W^i_{\alpha}\in \R^{r\times n}$ instead of $\W^i_{\beta}\in \R^{m\times r}$.

\begin{figure}[htbp]
    \centering
    \includegraphics[width=0.75\linewidth]{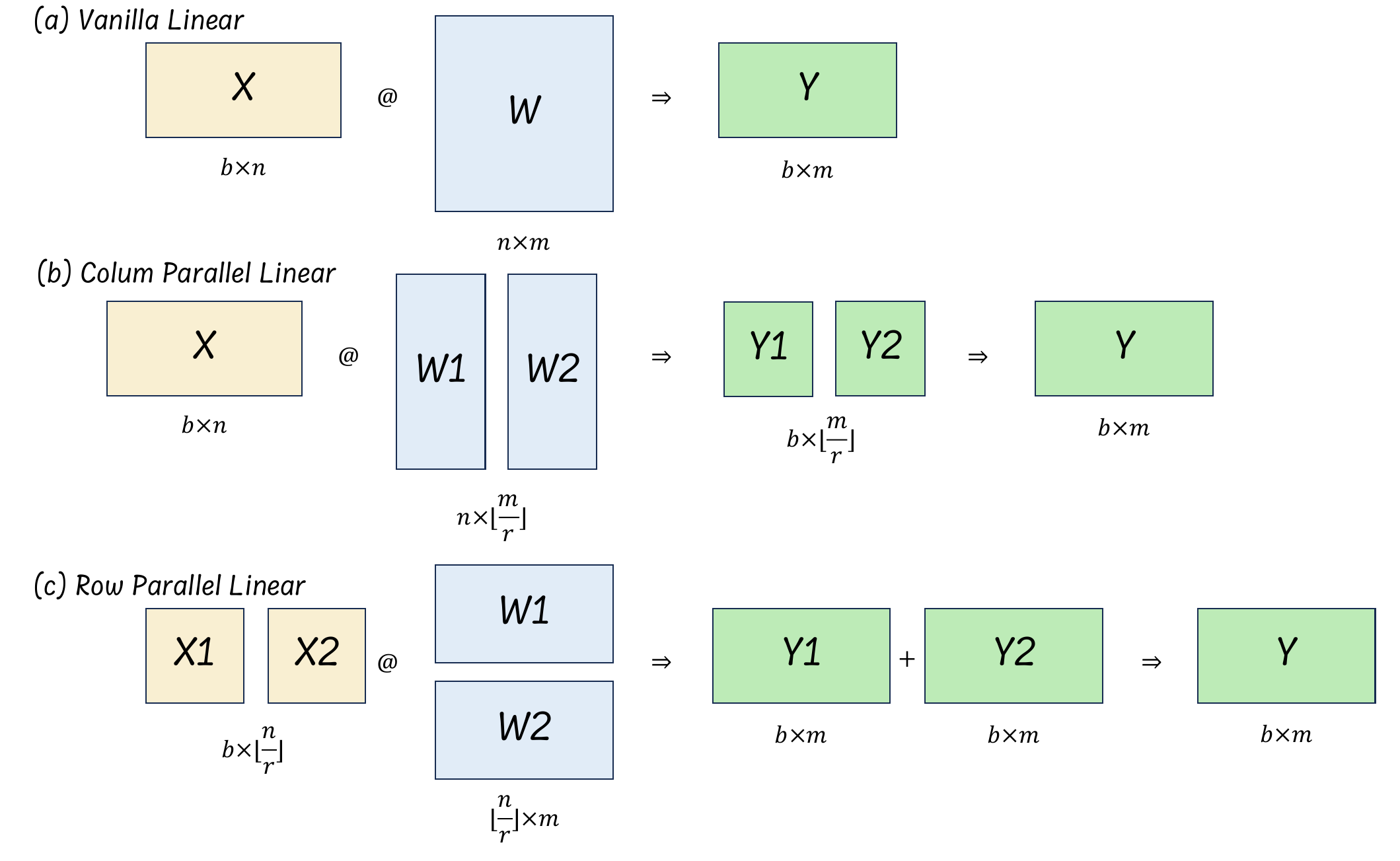}
    \caption{The vanilla linear function and the two parallel methods in Megatron-LM~\cite{shoeybi2019megatron}.}
    \label{fig:row_vs_parallel}
\end{figure}

In our paper, we set $\W^i_{\alpha}\in \R^{r\times n}$, thus we can utilize the memory optimization technique mentioned in Sec.~\ref{sec:method_proposed} to eliminate the storage of intermediate matrix $\text{Repeat}_0(\W^i_{\alpha}, \left\lfloor \frac{m}{r} \right\rfloor)$. The optimization approach follows the column parallel linear in Megatron-LM~\cite{shoeybi2019megatron}, corresponding to Fig.~\ref{fig:row_vs_parallel} (b). As can be seen, if we split the matrices from the dimension of $m$, we can separately calculate the linear operations $\Y_1=\F(\X,\W_1),\Y_2=\F(\X,\W_2),\dots,\Y_r=\F(\X,\W_{r})$ and concat them. Each $\Y_i$ ($1\leq i\leq r$) is a matrix of $\R^{b\times \left\lfloor \frac{m}{r} \right\rfloor}$. This does not lead to other additional intermediate weights. 

If we use $\W^i_{\beta}\in \R^{m\times r}$, the optimization method will correspond to row parallel linear as in Fig.~\ref{fig:row_vs_parallel} (c). This leads to two disadvantages: \textbf{(1)} We need to split both $\X$ and $\W$ (more steps to implement). \textbf{(2)} After applying $\Y_1=\F(\X_1,\W_1),\Y_2=\F(\X_2,\W_2),\dots,\Y_r=\F(\X_r,\W_r)$, we get intermediate matrices with size $b\times m$ (we have to store at least two) and need to add them together (less efficient).
\newpage
\subsection{Detailed Proof of Eq.~\ref{equ:matrix_memory}}\label{append:proof_of_matrix_mem}

Let us consider a more general case with $b\geq 1$. 

Given input activation $\X\in \R^{b\times n}$, $\W\in \R^{m\times n}$, $\W_{\alpha}\in \R^{r\times n}$ and $\W_{\beta}\in \R^{m\times 1}$, 

\begin{equation}
\begin{split}
&\X(\W\odot \text{Repeat}_0(\W_{\alpha}, \left\lfloor \frac{m}{r} \right\rfloor)\odot \text{Repeat}_1(\W_{\beta}, n))^T\\
=&\begin{bmatrix}
    x_{11}&x_{12}&\cdots&x_{1n}\\
    x_{21}&x_{22}&\cdots&x_{2n}\\
    \vdots&\vdots&&\vdots\\
    x_{b1}&x_{b2}&\cdots&x_{bn}\\
\end{bmatrix}\begin{bmatrix}
w_{11}\alpha_{11}\beta_1&w_{12}\alpha_{12}\beta_1&\cdots&w_{1n}\alpha_{1n}\beta_1\\
w_{21}\alpha_{11}\beta_2&w_{22}\alpha_{12}\beta_2&\cdots&w_{2n}\alpha_{1n}\beta_2\\
\vdots&\vdots&&\vdots\\
w_{\left\lfloor \frac{m}{r} \right\rfloor 1}\alpha_{11}\beta_{\left\lfloor \frac{m}{r} \right\rfloor}&w_{\left\lfloor \frac{m}{r} \right\rfloor 2}\alpha_{12}\beta_{\left\lfloor \frac{m}{r} \right\rfloor}&\cdots&w_{\left\lfloor \frac{m}{r} \right\rfloor n}\alpha_{1n}\beta_{\left\lfloor \frac{m}{r} \right\rfloor}\\
\vdots&\vdots&&\vdots\\
w_{m1}\alpha_{r1}\beta_m&w_{m2}\alpha_{r2}\beta_m&\cdots&w_{mn}\alpha_{rn}\beta_m\\
\end{bmatrix}^T\\ \nonumber
\end{split}
\end{equation}
\begin{equation}
\begin{split}
=&\begin{bmatrix}
    x_{11}&x_{12}&\cdots&x_{1n}\\
    x_{21}&x_{22}&\cdots&x_{2n}\\
    \vdots&\vdots&&\vdots\\
    x_{b1}&x_{b2}&\cdots&x_{bn}\\
\end{bmatrix}\begin{bmatrix}
w_{11}\alpha_{11}\beta_1&w_{21}\alpha_{11}\beta_2&\cdots&w_{\left\lfloor \frac{m}{r} \right\rfloor1}\alpha_{11}\beta_{\left\lfloor \frac{m}{r} \right\rfloor}&\cdots&w_{m1}\alpha_{r1}\beta_m\\
w_{12}\alpha_{12}\beta_1&w_{22}\alpha_{12}\beta_2&\cdots&w_{\left\lfloor \frac{m}{r} \right\rfloor2}\alpha_{12}\beta_{\left\lfloor \frac{m}{r} \right\rfloor}&\cdots&w_{m2}\alpha_{r2}\beta_m\\
\vdots&\vdots&&\vdots&&\vdots\\
w_{1n}\alpha_{1n}\beta_1&w_{2n}\alpha_{1n}\beta_2&\cdots&w_{\left\lfloor \frac{m}{r} \right\rfloor n}\alpha_{1n}\beta_{\left\lfloor \frac{m}{r} \right\rfloor}&\cdots&w_{mn}\alpha_{rn}\beta_m\\
\end{bmatrix}
\end{split}
\end{equation}

The matrix on the right can be split horizontally into $r$ blocks, each $\in \R^{n\times \left\lfloor \frac{m}{r} \right\rfloor}$. Let's take the first block for simplicity. Some notations should be added for clarity. $\W_{\alpha i}$ denotes the $i$-th row of $\W_{\alpha}$, $\W_i$ denotes the $i$-th block of $\W$ corresponding to the horizontal split, and $\W_i\in \R^{\left\lfloor \frac{m}{r} \right\rfloor \times n}$, $\W_{\beta i}$ denotes the $i$-th block of $\W_{\beta}$ corresponding to the horizontal split, and $\W_{\beta i}\in \R^{\left\lfloor \frac{m}{r} \right\rfloor \times 1}$.

\begin{equation}
\begin{split}
&\begin{bmatrix}
    x_{11}&x_{12}&\cdots&x_{1n}\\
    x_{21}&x_{22}&\cdots&x_{2n}\\
    \vdots&\vdots&&\vdots\\
    x_{b1}&x_{b2}&\cdots&x_{bn}\\
\end{bmatrix}\begin{bmatrix}
w_{11}\alpha_{11}\beta_1&w_{21}\alpha_{11}\beta_2&\cdots&w_{\left\lfloor \frac{m}{r} \right\rfloor1}\alpha_{11}\beta_{\left\lfloor \frac{m}{r} \right\rfloor}\\
w_{12}\alpha_{12}\beta_1&w_{22}\alpha_{12}\beta_2&\cdots&w_{\left\lfloor \frac{m}{r} \right\rfloor2}\alpha_{12}\beta_{\left\lfloor \frac{m}{r} \right\rfloor}\\
\vdots&\vdots&&\vdots\\
w_{1n}\alpha_{1n}\beta_1&w_{2n}\alpha_{1n}\beta_2&\cdots&w_{\left\lfloor \frac{m}{r} \right\rfloor n}\alpha_{1n}\beta_{\left\lfloor \frac{m}{r} \right\rfloor}\\
\end{bmatrix}\\
=&\begin{bmatrix}
    x_{11}\alpha_{11}&x_{12}\alpha_{12}&\cdots&x_{1n}\alpha_{1n}\\
    x_{21}\alpha_{11}&x_{22}\alpha_{12}&\cdots&x_{2n}\alpha_{1n}\\
    \vdots&\vdots&&\vdots\\
    x_{b1}\alpha_{11}&x_{b2}\alpha_{12}&\cdots&x_{bn}\alpha_{1n}\\
\end{bmatrix}\begin{bmatrix}
w_{11}\beta_1&w_{21}\beta_2&\cdots&w_{\left\lfloor \frac{m}{r} \right\rfloor1}\beta_{\left\lfloor \frac{m}{r} \right\rfloor}\\
w_{12}\beta_1&w_{22}\beta_2&\cdots&w_{\left\lfloor \frac{m}{r} \right\rfloor2}\beta_{\left\lfloor \frac{m}{r} \right\rfloor}\\
\vdots&\vdots&&\vdots\\
w_{1n}\beta_1&w_{2n}\beta_2&\cdots&w_{\left\lfloor \frac{m}{r} \right\rfloor n}\beta_{\left\lfloor \frac{m}{r} \right\rfloor}\\
\end{bmatrix}\\
=&\begin{bmatrix}
    x_{11}\alpha_{11}&x_{12}\alpha_{12}&\cdots&x_{1n}\alpha_{1n}\\
    x_{21}\alpha_{11}&x_{22}\alpha_{12}&\cdots&x_{2n}\alpha_{1n}\\
    \vdots&\vdots&&\vdots\\
    x_{b1}\alpha_{11}&x_{b2}\alpha_{12}&\cdots&x_{bn}\alpha_{1n}\\
\end{bmatrix}\begin{bmatrix}
w_{11}&w_{21}&\cdots&w_{\left\lfloor \frac{m}{r} \right\rfloor1}\\
w_{12}&w_{22}&\cdots&w_{\left\lfloor \frac{m}{r} \right\rfloor2}\\
\vdots&\vdots&&\vdots\\
w_{1n}&w_{2n}&\cdots&w_{\left\lfloor \frac{m}{r} \right\rfloor n}\\
\end{bmatrix}\odot \begin{bmatrix}
    \beta_1&\beta_2&\cdots&\beta_{\left\lfloor \frac{m}{r} \right\rfloor}\\
    \beta_1&\beta_2&\cdots&\beta_{\left\lfloor \frac{m}{r} \right\rfloor}\\
    \vdots&\vdots&&\vdots\\
    \beta_1&\beta_2&\cdots&\beta_{\left\lfloor \frac{m}{r} \right\rfloor}\\
\end{bmatrix}\\
=&\begin{bmatrix}
    x_{11}&x_{12}&\cdots&x_{1n}\\
    x_{21}&x_{22}&\cdots&x_{2n}\\
    \vdots&\vdots&&\vdots\\
    x_{b1}&x_{b2}&\cdots&x_{bn}\\
\end{bmatrix}\odot \begin{bmatrix}
    \alpha_{11}&\alpha_{12}&\cdots&\alpha_{1n}\\
    \alpha_{11}&\alpha_{12}&\cdots&\alpha_{1n}\\
    \vdots&\vdots&&\vdots\\
    \alpha_{11}&\alpha_{12}&\cdots&\alpha_{1n}\\
\end{bmatrix}\begin{bmatrix}
w_{11}&w_{21}&\cdots&w_{\left\lfloor \frac{m}{r} \right\rfloor1}\\
w_{12}&w_{22}&\cdots&w_{\left\lfloor \frac{m}{r} \right\rfloor2}\\
\vdots&\vdots&&\vdots\\
w_{1n}&w_{2n}&\cdots&w_{\left\lfloor \frac{m}{r} \right\rfloor n}\\
\end{bmatrix}\odot \begin{bmatrix}
    \beta_1&\beta_2&\cdots&\beta_{\left\lfloor \frac{m}{r} \right\rfloor}\\
    \beta_1&\beta_2&\cdots&\beta_{\left\lfloor \frac{m}{r} \right\rfloor}\\
    \vdots&\vdots&&\vdots\\
    \beta_1&\beta_2&\cdots&\beta_{\left\lfloor \frac{m}{r} \right\rfloor}\\
\end{bmatrix}\\
=&\X\odot \text{Repeat}_0(\W_{\alpha0},b)\W_1^T\odot \text{Repeat}_1(\W_{\beta0},b)^T\in \R^{b\times \left\lfloor \frac{m}{r} \right\rfloor}
\end{split}
\end{equation}

Similarly, the result of the $i$-th block is:
\begin{align}
\X\odot \text{Repeat}_0(\W_{\alpha i},b)\W_i^T\odot \text{Repeat}_1(\W_{\beta i},b)^T
\end{align}

Concatenating the result of all the $r$ blocks, we will get the final result:
\begin{equation}
\begin{split}
&\X(\W\odot \text{Repeat}_0(\W_{\alpha}, \left\lfloor \frac{m}{r} \right\rfloor)\odot \text{Repeat}_1(\W_{\beta}, n))^T\\
=&[\X\odot \text{Repeat}_0(\W_{\alpha 0},b)\W_0^T\odot \text{Repeat}_1(\W_{\beta 0},b)^T\cdots\X\odot \text{Repeat}_0(\W_{\alpha (r-1)},b)\W_{r-1}^T\odot \text{Repeat}_1(\W_{\beta (r-1)},b)^T]\\
=&[\X\odot \text{Repeat}_0(\W_{\alpha 0},b)\W_0^T\cdots\X\odot \text{Repeat}_0(\W_{\alpha (r-1)},b)\W_{r-1}^T]\odot [\text{Repeat}_1(\W_{\beta 0},b)^T\cdots \text{Repeat}_1(\W_{\beta (r-1)},b)^T]\\
=&[\X\odot \text{Repeat}_0(\W_{\alpha 0},b)\W_0^T\cdots\X\odot \text{Repeat}_0(\W_{\alpha (r-1)},b)\W_{r-1}^T]\odot \text{Repeat}_1(\W_{\beta},b)^T\\
=&[\cdots \X\odot \text{Repeat}_0(\W_{\alpha i},b)\W_i^T \cdots]\odot \text{Repeat}_1(\W_{\beta},b)^T
\end{split}
\end{equation}

If $b=1$, we get the same result with Eq.~\ref{equ:matrix_memory}, as $\text{Repeat}_d(\W,1)=\W$. Notice that $\W_{\alpha i}$ and $\W_{\beta}$ are all vectors, and PyTorch automatically aligns the matrix dimensions when doing Hadamard product between a
matrix and a vector. Besides, $\W_i$ is part of the frozen weights $\W$. During the training step, we do not need to save other redundant matrices.

% \subsection{Hyper Parameters}

% For training 7B/13B/30B/65B/70B models, we use learning rates of 4e-3/2e-3/4e-3/5e-4/5e-4 with per-device batch size set to 8/4/16/8/8. Following~\cite{dettmers2023qlora}, we set a 0.03 warm-up ratio, but decay the learning rate after reaching the targeted peak value. We use the AdamW optimizer with default setting in the Transformers package\footnote{https://github.com/huggingface/transformers} and add a 0.001 weight decay.

% After fixing all the hyper-parameters, we train post-training pruned LLMs obtained by SparseGPT~\cite{Frantar2023SparseGPTML} and Wanda~\cite{sun2023simple} with different sparsity patterns and ratios. We do not carry out hyper-parameter tuning for specific sparsity patterns or ratios. Experiment results shown in the paper demonstrate that our SPP can bring about stable performance improvements.
\newpage
\subsection{Zero-shot results for LLaMA-2}\label{sec:llama2_res}

% \usepackage{multirow}
% \usepackage{graphicx}
\begin{table}[h!]
\centering
\resizebox{\columnwidth}{!}{%
\begin{tabular}{clccccccccc}
\midrule
\textbf{LLaMA - 2} &
  \textbf{Method} &
  \textbf{Sparsity} &
  \textbf{BoolQ} &
  \textbf{RTE} &
  \textbf{HellaSwag} &
  \textbf{WinoGrande} &
  \textbf{ARC-e} &
  \textbf{ARC-c} &
  \textbf{OBQA} &
  \textbf{Average} \\ \midrule
\multirow{9}{*}{\textbf{7B}}  & None            & Dense            & 77.74 & 62.82 & 57.17 & 68.98 & 76.30 & 43.43 & 31.40 & 59.69 \\ \cmidrule{2-11} 
                              & SparseGPT      & Unstructured 50\% & 76.24 & 55.96 & 52.87 & 68.98 & 71.55 & 37.80 & 28.60 & 56.00 \\ %\cmidrule{2-11} 
                              & SparseGPT\textbf{+SPP} & Unstructured 50\% & 75.75 & 56.32 & 55.32 & 69.14 & 74.20 & 42.49 & 30.80 & 57.72 \\ %\cmidrule{2-11} 
                              & SparseGPT      & 2:4              & 66.91 & 57.76 & 43.55 & 66.14 & 64.10 & 31.40 & 24.20 & 50.58 \\ %\cmidrule{2-11} 
                              & SparseGPT\textbf{+SPP} & 2:4              & 70.89 & 54.15 & 51.33 & 67.40 & 70.96 & 38.99 & 29.00 & 54.67 \\ \cmidrule{2-11} 
                              & Wanda          & Unstructured 50\% & 76.73 & 53.43 & 52.50 & 68.43 & 72.35 & 39.16 & 30.80 & 56.20 \\ %\cmidrule{2-11} 
                              & Wanda\textbf{+SPP}     & Unstructured 50\% & 77.31 & 54.15 & 55.19 & 67.96 & 74.96 & 42.24 & 33.00 & 57.83 \\ %\cmidrule{2-11} 
                              & Wanda          & 2:4              & 68.47 & 53.07 & 41.33 & 62.67 & 62.88 & 30.55 & 23.60 & 48.94 \\ %\cmidrule{2-11} 
                              & Wanda\textbf{+SPP}     & 2:4              & 72.91 & 54.51 & 50.61 & 64.48 & 70.71 & 37.88 & 28.60 & 54.24 \\ \midrule
\multirow{9}{*}{\textbf{13B}} & None            & Dense            & 80.55 & 65.34 & 60.04 & 72.14 & 79.42 & 48.46 & 35.20 & 63.02 \\ \cmidrule{2-11} 
                              & SparseGPT      & Unstructured 50\% & 81.93 & 62.45 & 55.83 & 70.88 & 75.13 & 42.49 & 32.40 & 60.16 \\ %\cmidrule{2-11} 
                              & SparseGPT\textbf{+SPP} & Unstructured 50\% & 80.67 & 65.70 & 58.47 & 70.88 & 77.02 & 46.59 & 34.20 & 61.93 \\ %\cmidrule{2-11} 
                              & SparseGPT      & 2:4              & 78.62 & 56.68 & 48.32 & 68.51 & 68.52 & 36.01 & 27.40 & 54.87 \\ %\cmidrule{2-11} 
                              & SparseGPT\textbf{+SPP} & 2:4              & 78.10 & 67.15 & 55.29 & 70.24 & 72.73 & 43.00 & 30.80 & 59.62 \\ \cmidrule{2-11} 
                              & Wanda          & Unstructured 50\% & 81.13 & 58.84 & 57.07 & 70.72 & 75.67 & 42.83 & 32.20 & 59.78 \\ %\cmidrule{2-11} 
                              & Wanda\textbf{+SPP}     & Unstructured 50\% & 80.92 & 69.31 & 58.45 & 71.82 & 78.49 & 46.84 & 34.20 & 62.86 \\ %\cmidrule{2-11} 
                              & Wanda          & 2:4              & 76.09 & 55.96 & 46.19 & 67.56 & 68.64 & 34.13 & 24.20 & 53.25 \\ %\cmidrule{2-11} 
                              & Wanda\textbf{+SPP}     & 2:4              & 76.70 & 66.43 & 55.02 & 68.11 & 72.60 & 42.24 & 31.00 & 58.87 \\ \midrule
\multirow{9}{*}{\textbf{70B}} & None            & Dense            & 83.45 & 67.87 & 66.05 & 77.98 & 82.55 & 54.44 & 37.20 & 67.08 \\ \cmidrule{2-11} 
                              & SparseGPT      & Unstructured 50\% & 84.75 & 71.84 & 64.10 & 78.22 & 81.55 & 52.73 & 37.20 & 67.20 \\ %\cmidrule{2-11} 
                              & SparseGPT\textbf{+SPP} & Unstructured 50\% & 84.75 & 72.20 & 64.05 & 78.06 & 81.65 & 52.73 & 37.20 & 67.23 \\ %\cmidrule{2-11} 
                              & SparseGPT      & 2:4              & 81.30 & 69.31 & 58.60 & 76.24 & 79.50 & 48.21 & 32.60 & 63.68 \\ %\cmidrule{2-11} 
                              & SparseGPT\textbf{+SPP} & 2:4              & 81.30 & 68.59 & 58.80 & 76.24 & 79.40 & 48.29 & 32.60 & 63.60 \\ \cmidrule{2-11} 
                              & Wanda          & Unstructured 50\% & 82.50 & 72.56 & 63.90 & 78.06 & 81.60 & 52.47 & 37.80 & 66.98 \\ %\cmidrule{2-11} 
                              & Wanda\textbf{+SPP}     & Unstructured 50\% & 82.25 & 72.92 & 64.40 & 77.90 & 81.70 & 53.24 & 37.60 & 67.14 \\ %\cmidrule{2-11} 
                              & Wanda          & 2:4              & 79.55 & 67.87 & 59.05 & 76.24 & 79.40 & 47.70 & 35.40 & 63.60 \\ %\cmidrule{2-11} 
                              & Wanda\textbf{+SPP}     & 2:4              & 80.25 & 69.31 & 60.65 & 76.01 & 80.10 & 48.81 & 35.20 & 64.33 \\ \midrule
\end{tabular}%
}
\caption{Zero-shot evaluation of 7 different tasks from EleutherAI LM Harness~\cite{gao2021framework} after training LLaMA-2 on Stanford-Alpaca~\cite{taori2023alpaca} by SPP. As can be seen from the table, SPP can improve the performance of sparse models.}
\label{tab:zero_llama2_50}
\end{table}

Here we provide in Tab.~\ref{tab:zero_llama2_50} the zero-shot performance on seven tasks from EleutherAI LM Harness~\cite{gao2021framework} after training the LLaMA-2 model family on the Alpaca dataset. We find that after instruction fine-tuning, 11 out of 12 models show performance improvement.
\newpage
\subsection{More Ablations}

More ablations concerning zero-initialization of added weight are provided. We carry out experiments with 4 different settings on the LLaMA-7B model (none zero-init, $\W_\alpha$ zero-init, $\W_\beta$ zero-init, and both zero-init). We observe that zero-initializing both $\W_\alpha$ and $\W_\beta$ will lead to gradient vanishment, and the added parameters will remain 0 all the time, so it will not bring any performance enhancement. Average zero-shot accuracies of the other three initialization choices on 7 tasks of LM-eval~\cite{gao2021framework} are shown in Tab.~\ref{tab:more_abla}.

% \usepackage{multirow}
% \usepackage{graphicx}
\begin{table}[h!]
\centering
\resizebox{0.55\columnwidth}{!}{%
\begin{tabular}{ccccc}
\midrule
{\color[HTML]{000000} \textbf{Method}} &
  {\color[HTML]{000000} \textbf{Sparsity}} &
  {\color[HTML]{000000} \textbf{Zero-init $\W_\alpha$}} &
  {\color[HTML]{000000} \textbf{Zero-init $\W_\beta$}} &
  {\color[HTML]{000000} \textbf{LM-eval}} \\ \midrule
{\color[HTML]{000000} } &
  {\color[HTML]{000000} } &
  {\color[HTML]{000000} } &
  {\color[HTML]{000000} } &
  {\color[HTML]{000000} 53.52} \\ % \cmidrule{3-5} 
{\color[HTML]{000000} } &
  {\color[HTML]{000000} } &
  {\color[HTML]{000000} \checkmark} &
  {\color[HTML]{000000} } &
  {\color[HTML]{000000} \textbf{56.16}} \\ % \cmidrule{3-5} 
{\color[HTML]{000000} } &
  \multirow{-3}{*}{{\color[HTML]{000000} 2:4}} &
  {\color[HTML]{000000} } &
  {\color[HTML]{000000} \checkmark} &
  {\color[HTML]{000000} 55.42} \\ \cmidrule{2-5} 
{\color[HTML]{000000} } &
  {\color[HTML]{000000} } &
  {\color[HTML]{000000} } &
  {\color[HTML]{000000} } &
  {\color[HTML]{000000} 57.59} \\ % \cmidrule{3-5} 
{\color[HTML]{000000} } &
  {\color[HTML]{000000} } &
  {\color[HTML]{000000} \checkmark} &
  {\color[HTML]{000000} } &
  {\color[HTML]{000000} 57.09} \\ % \cmidrule{3-5} 
\multirow{-6}{*}{{\color[HTML]{000000} {Wanda+\textbf{SPP}}}} &
  \multirow{-3}{*}{{\color[HTML]{000000} Unstructured 50\%}} &
  {\color[HTML]{000000} } &
  {\color[HTML]{000000} \checkmark} &
  {\color[HTML]{000000} \textbf{58.13}} \\ \midrule
{\color[HTML]{000000} } &
  {\color[HTML]{000000} } &
  {\color[HTML]{000000} } &
  {\color[HTML]{000000} } &
  {\color[HTML]{000000} 54.01} \\ % \cmidrule{3-5} 
{\color[HTML]{000000} } &
  {\color[HTML]{000000} } &
  {\color[HTML]{000000} \checkmark} &
  {\color[HTML]{000000} } &
  {\color[HTML]{000000} 53.95} \\ % \cmidrule{3-5} 
{\color[HTML]{000000} } &
  \multirow{-3}{*}{{\color[HTML]{000000} 2:4}} &
  {\color[HTML]{000000} } &
  {\color[HTML]{000000} \checkmark} &
  {\color[HTML]{000000} \textbf{54.60}} \\ \cmidrule{2-5} 
{\color[HTML]{000000} } &
  {\color[HTML]{000000} } &
  {\color[HTML]{000000} } &
  {\color[HTML]{000000} } &
  {\color[HTML]{000000} 57.12} \\ % \cmidrule{3-5} 
{\color[HTML]{000000} } &
  {\color[HTML]{000000} } &
  {\color[HTML]{000000} \checkmark} &
  {\color[HTML]{000000} } &
  {\color[HTML]{000000} 57.59} \\ % \cmidrule{3-5} 
\multirow{-6}{*}{{\color[HTML]{000000} {SparseGPT+\textbf{SPP}}}} &
  \multirow{-3}{*}{{\color[HTML]{000000} Unstructured 50\%}} &
  {\color[HTML]{000000} } &
  {\color[HTML]{000000} \checkmark} &
  {\color[HTML]{000000} \textbf{58.43}} \\ \midrule
\end{tabular}%
}
\caption{More ablation studies concerning zero-initialization of added weights. Zero-initializing $\W_\beta$ will get the overall best results in our experiments.}
\label{tab:more_abla}
\end{table}

As shown in the table, we find initializing $\W_\beta$ with 0 and randomly initializing $\W_\alpha$ lead to overall better results in our experiment setting.

\end{document}